%% file: main.tex
\documentclass{article}

\input{math_commands.tex}

\usepackage{multicol}
\usepackage{url}
\usepackage[colorlinks,linkcolor=blue,citecolor=blue, pagebackref=true,backref=true]{hyperref}
\usepackage{multirow}
\usepackage{sidecap}
\usepackage{cancel}
\usepackage{graphicx}
\usepackage{amsmath,amssymb}
\usepackage{float}
\usepackage[normalem]{ulem}

\usepackage[final]{neurips_distshift_2021}

\usepackage[utf8]{inputenc} 
\usepackage[T1]{fontenc}    
\usepackage{hyperref}       
\usepackage{url}            
\usepackage{booktabs}       
\usepackage{amsfonts}       
\usepackage{nicefrac}       
\usepackage{microtype}      
\usepackage{xcolor}         

\newcommand\blfootnote[1]{%
	\begingroup
	\renewcommand\thefootnote{}\footnote{#1}%
	\addtocounter{footnote}{-1}%
	\endgroup
}

\title{The Effect of Model Size on Worst-Group Generalization}

\author{Alan Pham$^*$,\, Eunice Chan$^*$,\, Vikranth Srivatsa$^*$,\, Dhruba Ghosh$^*$,  \\
\textbf{Yaoqing Yang,\, Yaodong Yu,\,  Ruiqi Zhong,\, } 
\textbf{Joseph E. Gonzalez$^\dagger$,\, Jacob Steinhardt$^\dagger$} \\
Department of Electrical Engineering and Computer Science \\
University of California, Berkeley\\
\texttt{\{alanlp, ekchan, vsrivatsa, djghosh13, yqyang, yaodong\_yu,} \\
\texttt{ruiqi\-zhong, jegonzal, jsteinhardt\}@berkeley.edu} \\}

\renewcommand{\paragraph}[1]{\textbf{#1}}

\begin{document}

\maketitle

\blfootnote{\hspace*{-3.4mm} Both $^*$ and $^\dagger$ mean equal contribution.}

\input{parts/0_Abstract.tex}

\input{parts/1_Intro.tex}

\input{parts/2_Related_works.tex}

\input{parts/3_Setup}


\input{parts/4_Experiment.tex}


\input{parts/5_Conclusion.tex}


\bibliographystyle{iclr2022_conference}
\bibliography{ref}

\appendix
\newpage
\section{Appendix}\label{Appendix}
\input{parts/7_Appendix.tex}

\end{document}

%% file: math_commands.tex
\usepackage{amsmath,amsfonts,bm}

\def\eqref#1{equation~\ref{#1}}

\def\1{\bm{1}}

\DeclareMathAlphabet{\mathsfit}{\encodingdefault}{\sfdefault}{m}{sl}
\SetMathAlphabet{\mathsfit}{bold}{\encodingdefault}{\sfdefault}{bx}{n}



%% file: parts/0_Abstract.tex
\begin{abstract}\label{abstract}

Overparameterization is shown to result in poor test accuracy on rare subgroups under a variety of settings where subgroup information is known. To gain a more complete picture, we consider the case where subgroup information is unknown. %
We investigate the effect of model size on worst-group generalization under empirical risk minimization (ERM) across a wide range of settings, varying: 1) architectures (ResNet, VGG, or BERT), 2) domains (vision or natural language processing), 3) model size (width or depth), and 4) initialization (with pre-trained or random weights).
Our systematic evaluation %
reveals that increasing model size does not hurt, and may help, worst-group test performance under ERM across all setups.
In particular, increasing pre-trained model size consistently improves performance on Waterbirds and MultiNLI. We advise practitioners to use larger pre-trained models when subgroup labels are unknown.

\end{abstract}

%% file: parts/1_Intro.tex
\section{Introduction}\label{introduction}

Recent work shows that overparameterized models display stronger generalization performance than smaller models, even on small datasets, suggesting that larger models overfit less \citep{bornschein2020small}. %
Despite this trend, many works find or assume that overparameterization can hurt test accuracy on certain subgroups of the data \citep{pmlr-v81-buolamwini18a, pmlr-v80-hashimoto18a, Menon2021OverparameterisationAW, distributionally, zhong2021larger}. 
This problem is exacerbated when tuning a model with unknown subgroup labels %
since the performance gap between average accuracy and worst-group accuracy--the subgroup with poorest accuracy-- can be large. The difference between the average test accuracy and worst-group test accuracy ranges from 40 to 60\% in the empirical risk minimization (ERM) setup from \citet{Sagawa2020AnIO} on CelebA and Waterbirds. In addition to the resulting fairness concerns, real-world deployments might encounter distribution shifts that upweight rare subgroups, causing the average accuracy to suffer.

Low worst-group accuracy often occurs when subgroups are associated with spurious features.
For example, consider the task of classifying images of landbirds 
and waterbirds: although waterbirds are more likely to appear on a water background, the image background (the spurious feature) has no direct causal relationship with the species of bird.
These spurious features harm performance on ``rare'' subgroups (e.g.~waterbirds on land background) where the cue contradicts the true label.
This can lead to high-stakes real-world problems: thoracic pathology detection models were found to rely on the presence of chest drains (a treatment device) to detect pneumothorax, which led to poor performance on the clinically relevant rare subgroup of untreated patients with pneumothorax \citep{hidden_strat_clinically_meaningful}.

We systematically investigate the effect of model size on the performance of the model on the rare subgroups.  
Prior works show that increasing model size can hurt worst-group performance. In \citet{Sagawa2020AnIO}, this trend was shown for models trained with the reweighted objective that upweights minority groups. Furthermore, \citet{Sagawa2020AnIO} finds that, trained with naive ERM, models have poor worst-group error regardless of model size. In our work, we provide complementary findings as we explore the trend in naive ERM models in more comprehensive and informative experiments.
We experiment on a wide range of datasets with spurious correlations, which include: Waterbirds (a dataset of birds on land and water) \citep{distributionally}, CelebA (a face dataset we use for two tasks: Lipstick / Earring and Blond / Male) \citep{liu2015faceattributes}, and MultiNLI (a language dataset) \citep{mnli2018} datasets (more details and conventions listed in Appendix \ref{experimental_setup}). 
We also conduct experiments across multiple model architectures, varying model depth or width and initializing with pre-trained or random weights.

We find that, under ERM, increasing model size does not hurt and sometimes helps worst-group test accuracy across all settings considered in this paper (Section \ref{experimental_results}). 
In particular, larger pre-trained models are often less susceptible to spurious correlation. 
For example, in the Waterbirds dataset, compared to pre-trained ResNet18, pre-trained ResNet152 decreases the worst-group error by approximately 18.39\% (Figure \ref{fig:pretrain-depth}).  

Interestingly, pre-trained models actually achieve \emph{better} worst-group accuracy as model size increases, while models trained from scratch merely do not get worse. %

We summarize our contributions as follows:
\begin{itemize}
\item We empirically show that, under ERM, larger model sizes either help or do not hurt worst-group test accuracy across a wide range of settings. Specifically, we look across vision and language datasets and common model architectures.
\item We explore the trend over increasing numbers of parameters by separately looking at the effects widening models and deepening models. We also look at the effects of pre-training the model on a different dataset compared to randomly initialized models.
\item We find that larger pre-trained models consistently improve worst-case group accuracies on two widely used datasets: Waterbirds and MultiNLI.
\end{itemize}

%% file: parts/2_Related_works.tex
\section{Related Work}\label{related_work}

\textbf{Generalization with Overparameterized Models.}
Many widely-used models, including the ResNet, VGG batchnorm (BN), and BERT series of architectures, show a monotonic increase in average test accuracy on common tasks with increasing model size. Overparameterization is also found to improve robustness to adversarial samples and distributional shifts \citep{hendrycks2021natural}. This has been attributed to the ``double descent'' phenomenon, which explains the increase in accuracy from significantly overparameterizing models \citep{belkin2019reconciling}. \citet{bornschein2020small} discusses this in the context of small data, while \citet{yang2020rethinking} provides an explanation via the bias-variance trade-off curve. \citet{Sagawa2020AnIO} studies overparameterization and worst-group accuracy, finding that larger models often fail to generalize to rare groups in the test distribution. This is attributed to the presence of spurious correlations in the training data.

\textbf{Distributional Shift and Spurious Correlation.}
Pre-trained language models are more robust against spurious correlation, though this is dependent on the number of negative samples (i.e., samples that for which the spurious correlation does not correctly predict the label) \citep{tu2020empirical}. On the other hand, \citet{probingbert2019} finds that language models heavily rely on spurious correlations. For instance, on the argument reasoning comprehension task, even when logically necessary parts of the input are obscured, BERT Large is still reliably able to perform well.
Meanwhile, vision models suffer from spurious correlations %
due to scene biases such as co-occurrence of objects with other objects, backgrounds, or textures \citep{zhou2021examining, hermann2020origins}.

\textbf{Mitigating Spurious Correlation.}
Prior works have explored data augmentation to combat dependence on spurious correlations. In NLP, one approach is counterfactual augmentation: modifying the input sentences by a minimal amount to change the target label \citep{counterfactual2019}. This reduces the strength of spurious correlations by introducing variation in core features (with a causal relation to the label) while controlling for non-causal variables. In computer vision, \citet{goel2020model} suggests using CycleGAN to generate negative examples.
\citet{liu2021just} suggests a two-stage training regime that upweights misclassified points. \citet{wang2021causal} introduces a causal attention module that performs unsupervised annotations to mitigate spurious correlation.

Alternatively, worst-group test error can be reduced through the use of distributionally robust optimization (DRO) to guide training \citep{zhou2021examining, hashimoto2018fairness}. One approach is group DRO (GDRO), an instance of DRO that
minimizes the worst-group expected loss %
\citep[Eqn.(4)]{distributionally}. This substantially improves generalization when coupled with heavy regularization, but requires prior labeling of subgroups in the training set.

%% file: parts/3_Setup.tex
\section{Problem Setting}\label{setup}

We adopt the formulation presented in \citet{Sagawa2020AnIO}. Each sample consists of an input $x \in \mathcal{X}$, a target label $y \in \mathcal{Y}$, and a spurious attribute $a \in \mathcal{A}$. 
 We categorize samples into groups $g = (y, a) \in \mathcal{G} = \mathcal{Y} \times \mathcal{A}$. The spurious attribute $a$ is correlated with the label $y$, but has no causal relationship. The problem domains we consider are all classification tasks, where $|\mathcal{Y}| = 2 \text{ or } 3$, and are confounded by a binary spurious attribute ($|\mathcal{A}| = 2$).
For example, in the Waterbirds dataset, we might have an image $x$ of a waterbird on a land background: the label $y$ is the target class ``waterbird'', the spurious feature $a$ is ``land background'', and the group $g$ is (waterbird, land background).

Our work focuses on the effect of model size on worst-group test error. %
We train using empirical risk minimization (ERM)~\citep{vapnik1998statistical}, finding the model parameters $\theta$ that empirically minimize the average training loss:
\begin{equation}
\mathcal{R}_{\text{ERM}}(\theta) = \mathbb{E}_{(x,y,g)}[\ell(\theta, (x, y))].
\end{equation}
We use cross-entropy loss and choose the weight decay, learning rate, and training epoch so that the models are trained to convergence.
Further details on the training procedure are in Appendix \ref{training_procedure}.

\paragraph{Metrics.} Given a model $h: \mathcal{X} \to \mathcal{Y}$, we define the error on a group $g \in \mathcal{G}$ as
\begin{equation}
    \varepsilon_g := \mathbb{E}_{x,y|g}[\mathbf{1}(h(x) \neq y)].
\end{equation} 
Throughout the paper, we compute the group error averaged over the last 10 epochs of training in order to reduce noise and smooth out the results.

We consider two metrics: the average error as well as the worst-group error, defined as:
\begin{align}
    \varepsilon := \sum_{g \in G} w_g \cdot \varepsilon_g \quad \text{ and } \quad
    \varepsilon_{\text{wg}} := \max_{g \in \mathcal{G}} \varepsilon_g,
\end{align}
where $w_g$ is the weight equal to the group's proportion in the training data.

Across our experiments on increasing model depths and widths for the ResNet, VGG BN, MobileNet, and BERT architectures, we report average and worst-group error on the training and test datasets.

%% file: parts/4_Experiment.tex
\section{Experimental Results}\label{experimental_results}

\begin{figure}
\noindent
\begin{minipage}[H]{0.24\linewidth}
\raggedleft
\includegraphics[scale=0.18]
{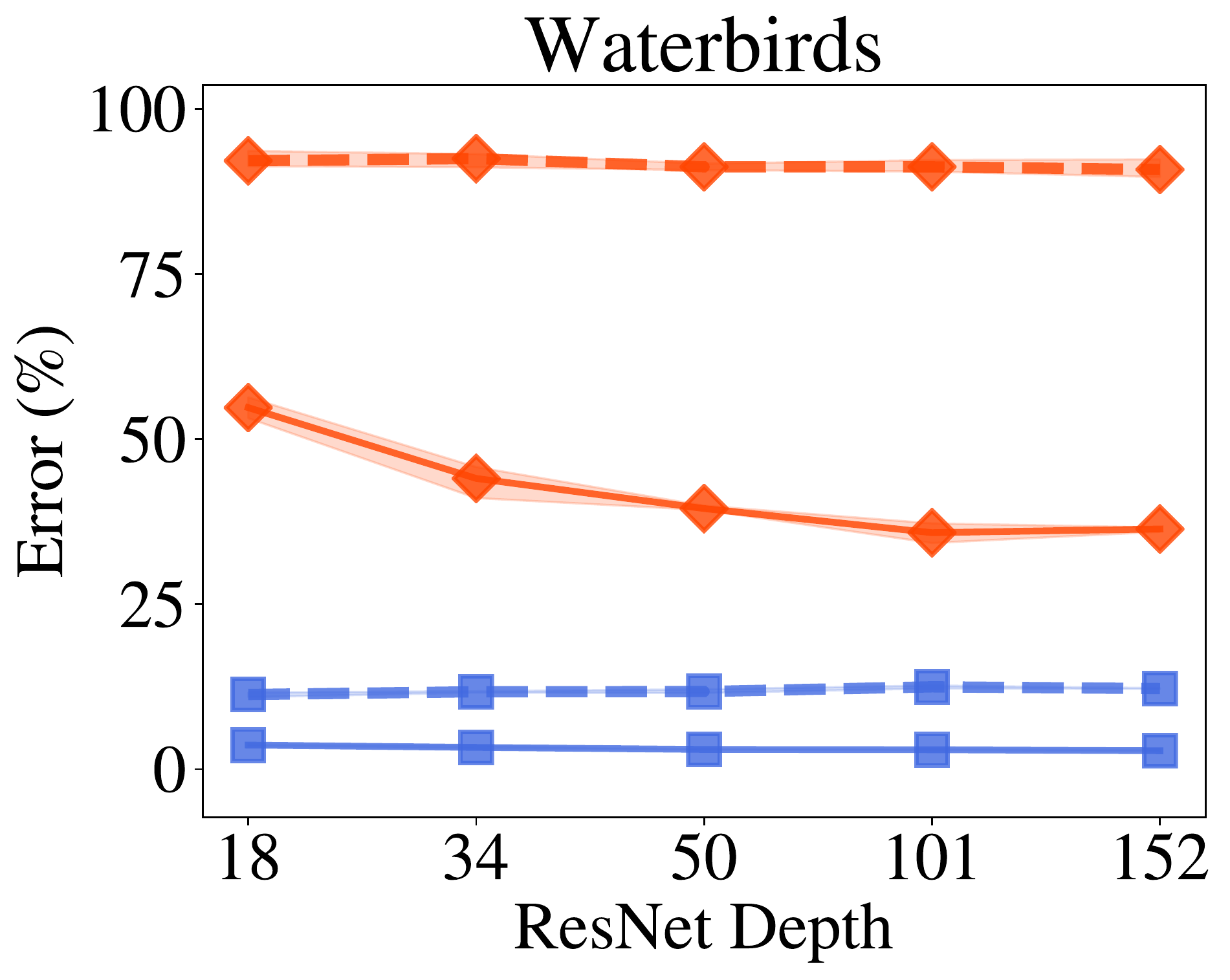}\\
\raggedleft
\includegraphics[scale=0.18]{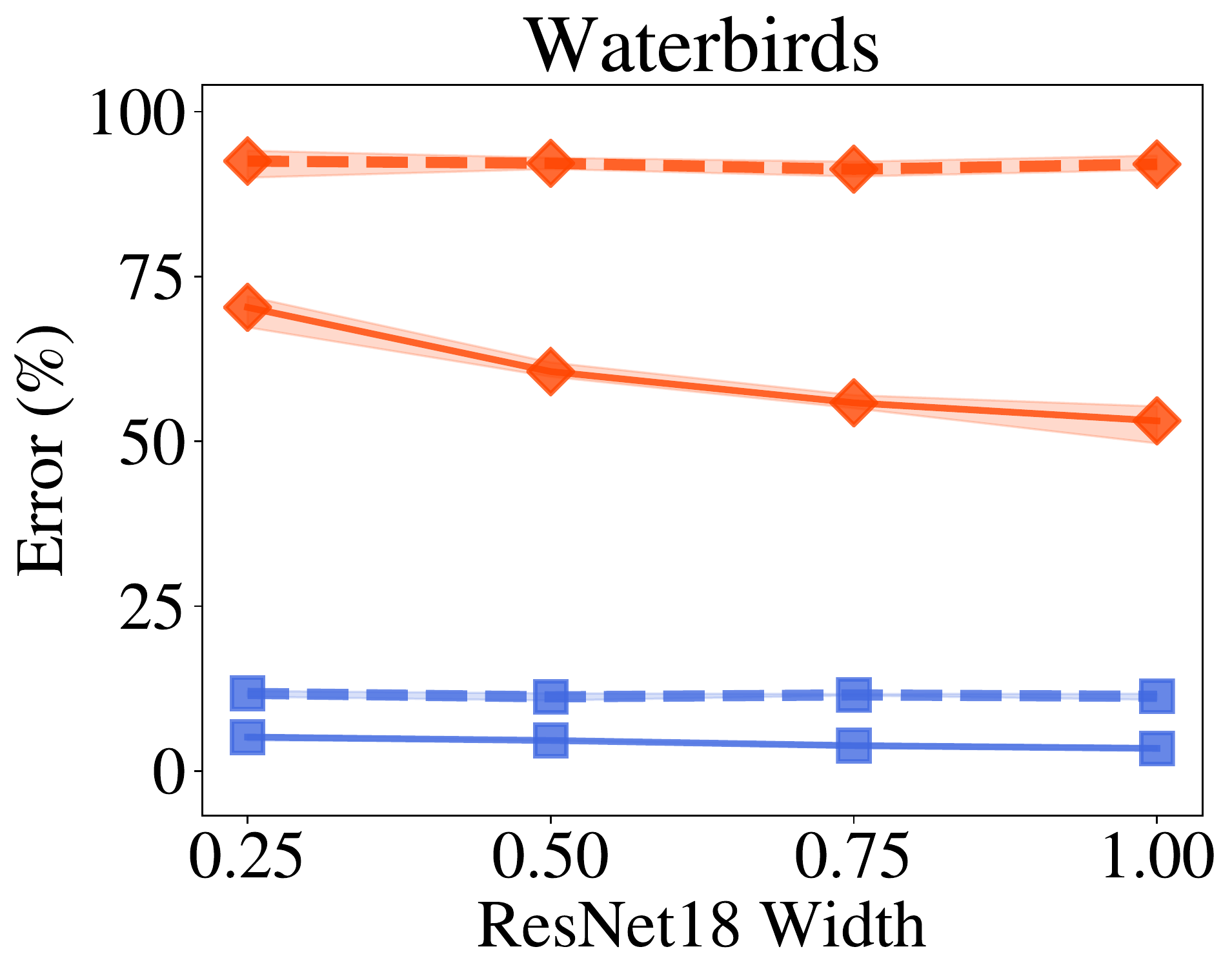}
\end{minipage}
\begin{minipage}[H]{0.24\linewidth}
\includegraphics[scale=0.18]{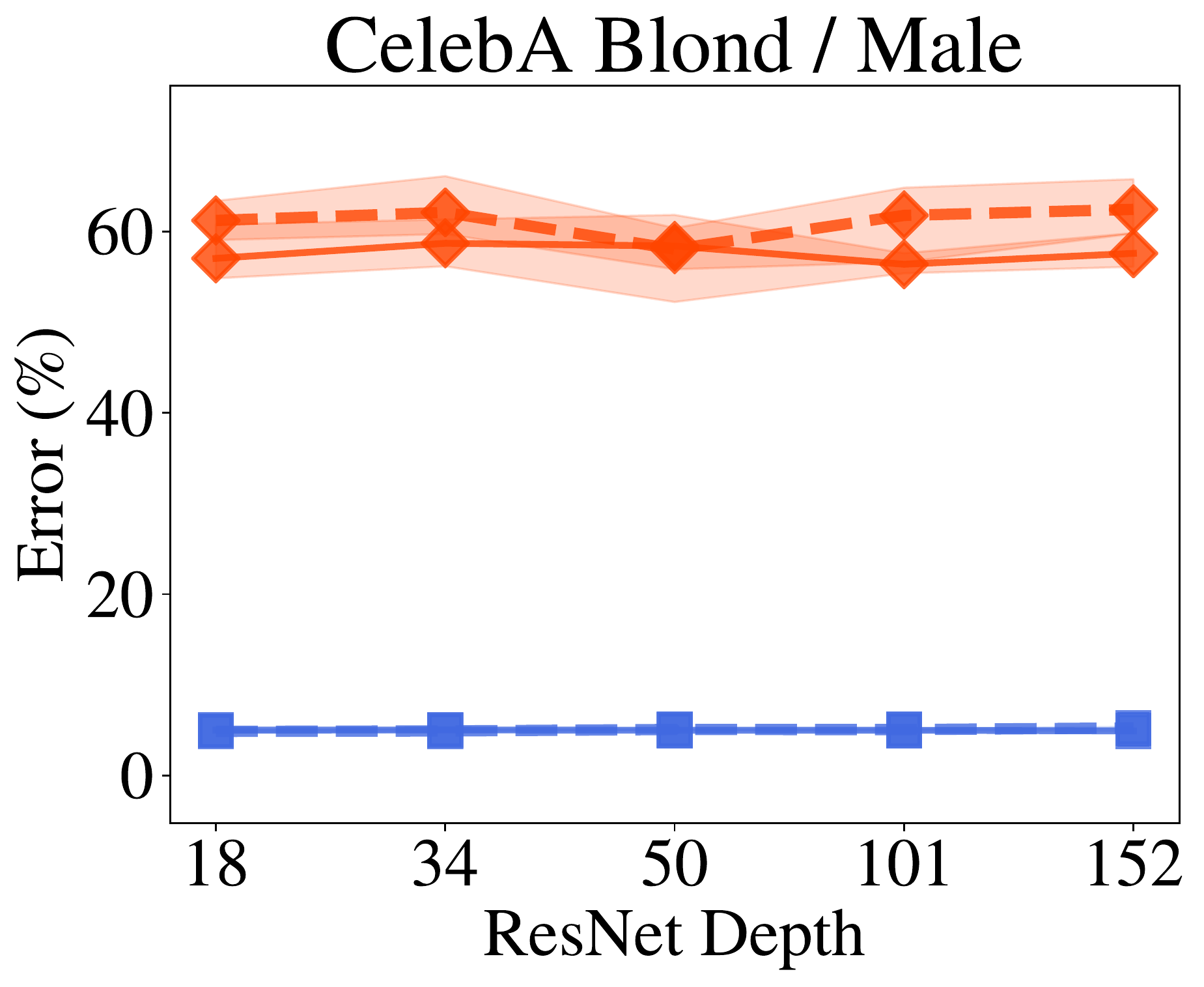}
\\
\includegraphics[scale=0.18]{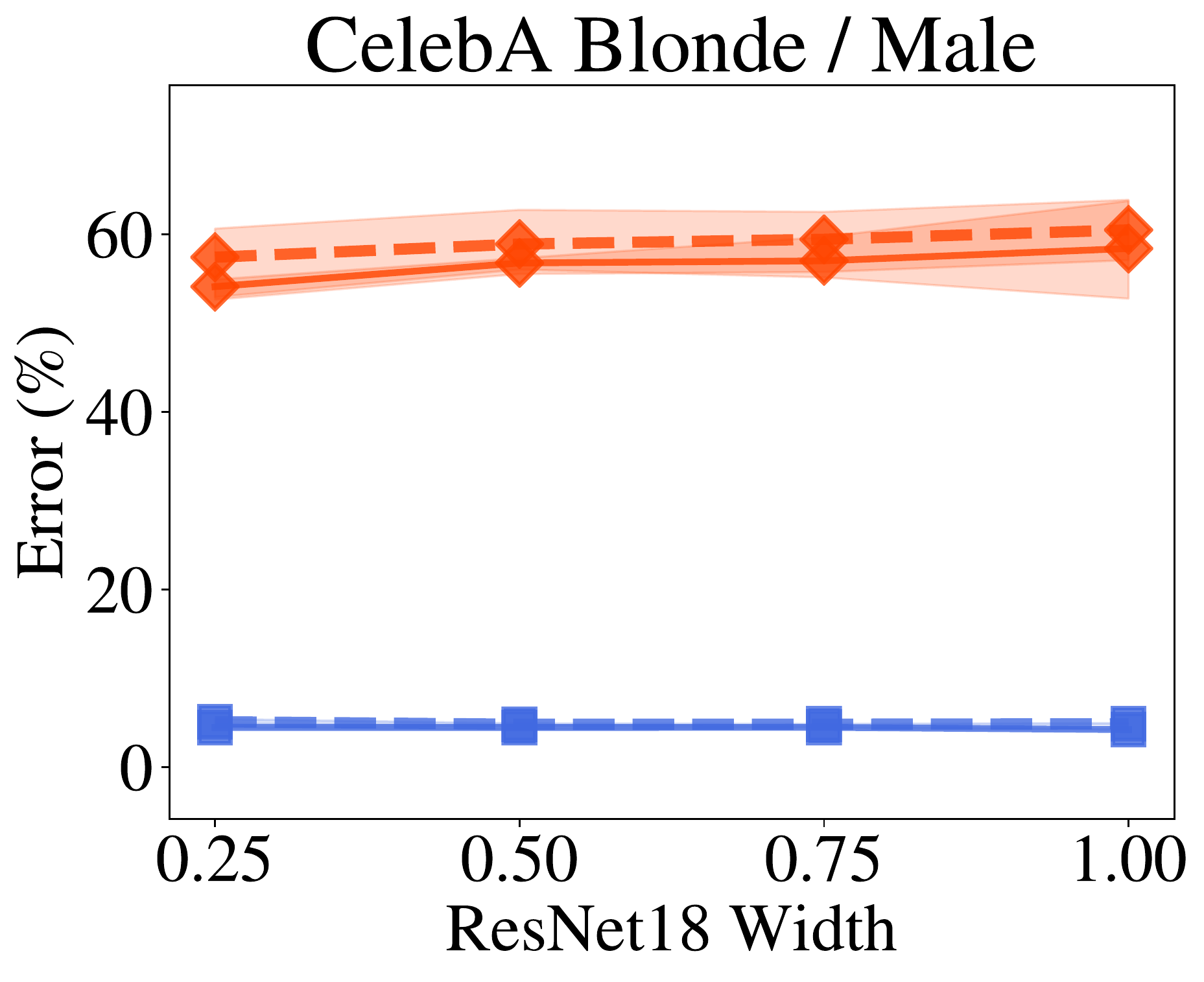}
\end{minipage}
\begin{minipage}[H]{0.24\linewidth}
\includegraphics[scale=0.18]{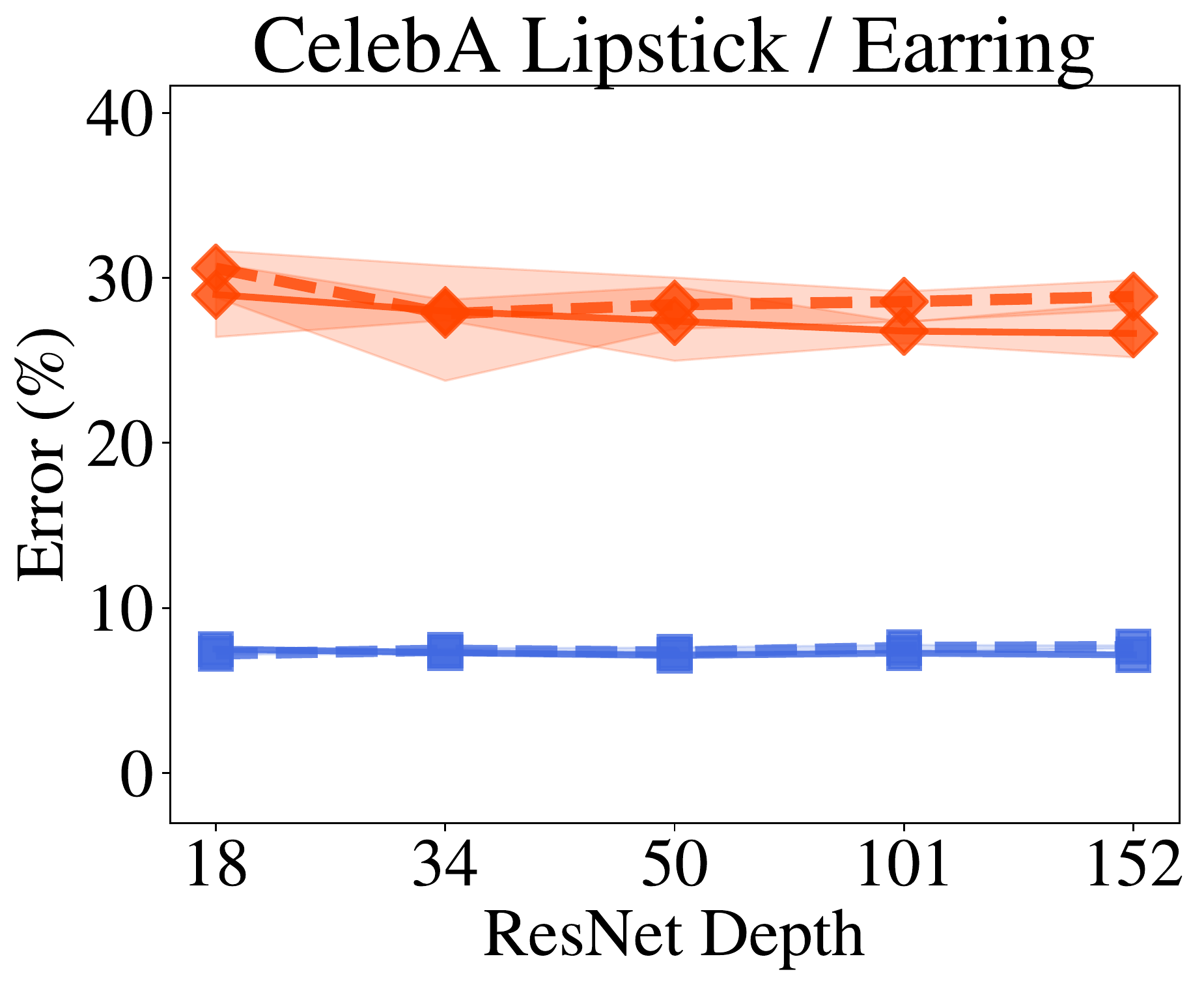}
\\
\raggedright
\includegraphics[scale=0.18]{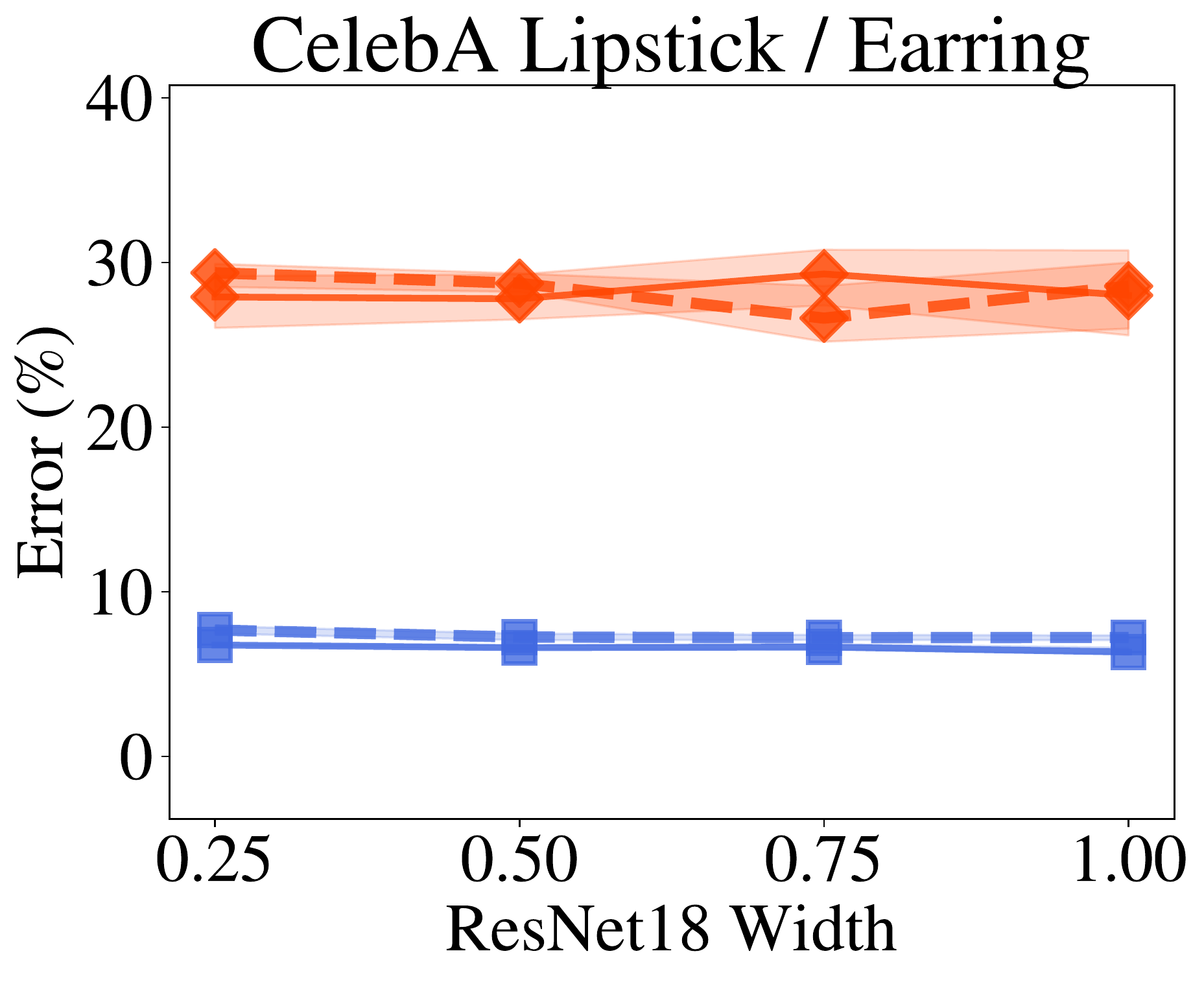}
\end{minipage}
\begin{minipage}[H]{0.24\linewidth}
\includegraphics[scale=0.18]{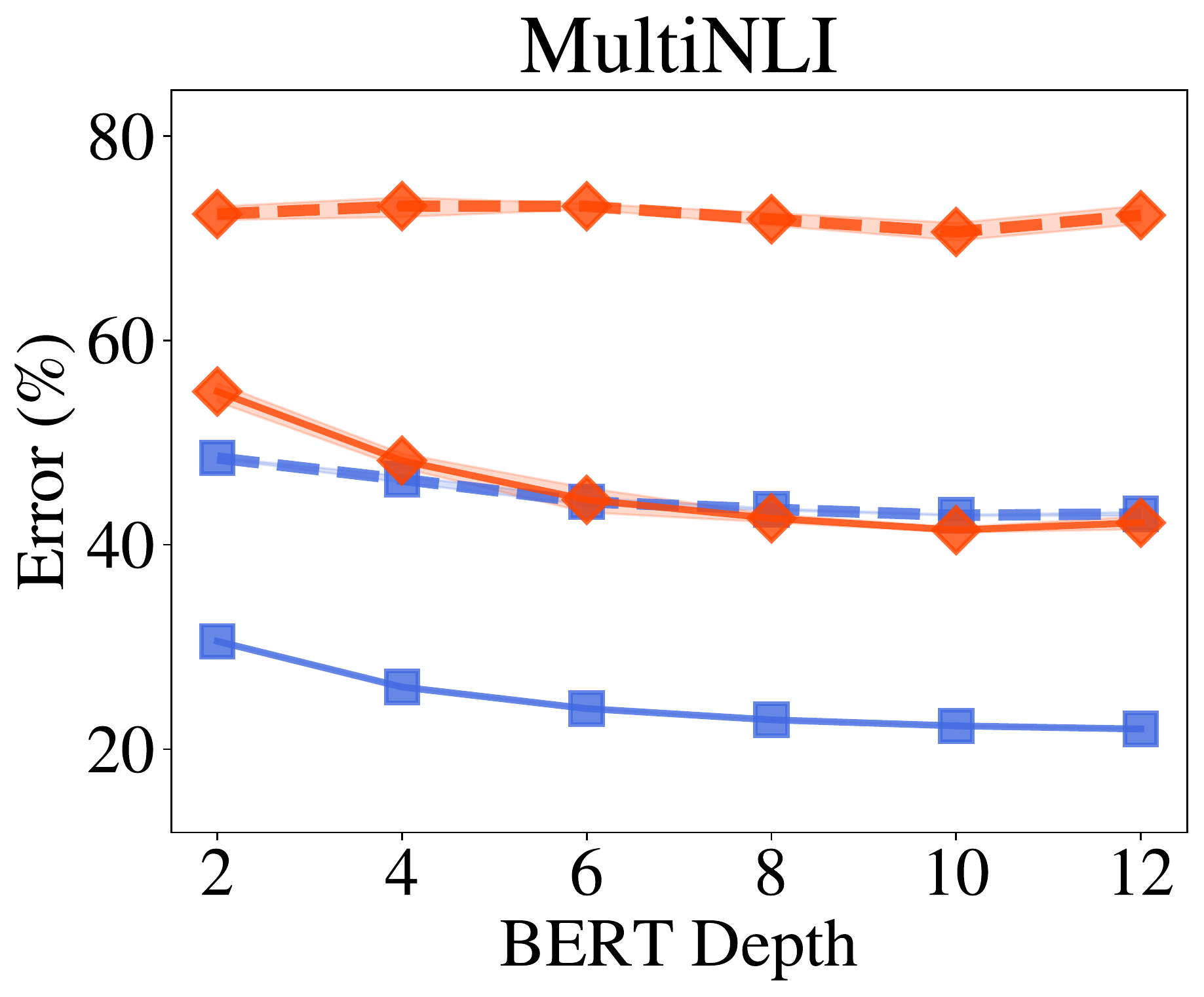}
\\
\includegraphics[scale=0.18]{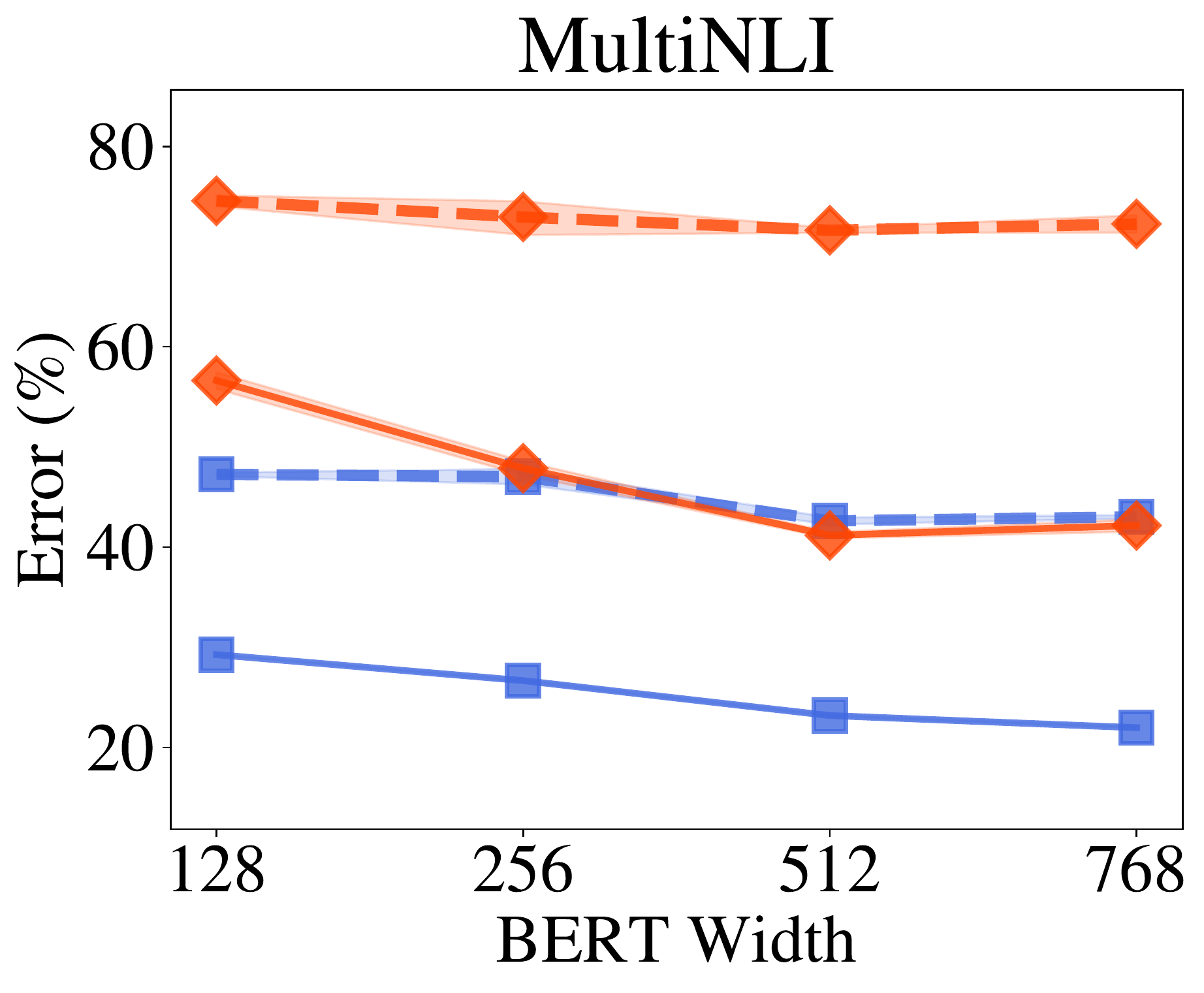}
\end{minipage}
\\
\begin{minipage}[H]{1\linewidth}
\center
\includegraphics[scale=0.15]{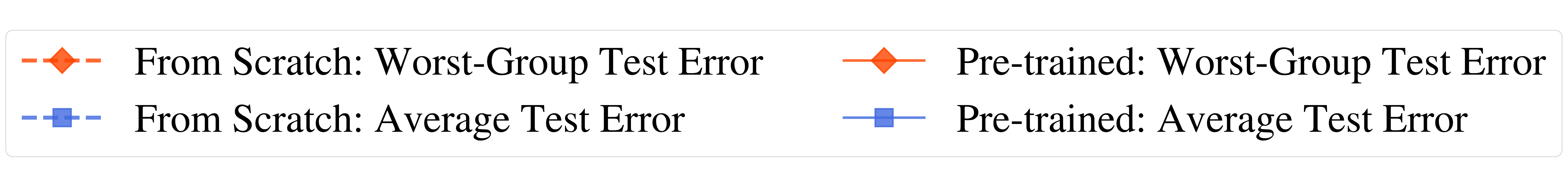}
\center
\end{minipage}
\caption{
Models trained to convergence, e.g., until average training accuracy is close or equal to 100\%.
Hyperparameters remain the same within each experiment series. In each of the four graphs above, we compare the error of pre-trained and randomly initialized models of the same architecture. Pre-trained models perform better than models trained from scratch and increasing model sizes for both types of models does not hurt the worst-group error. \textbf{Top row:} Depth-varying results. \textbf{Bottom row:} Width-varying results. \textbf{Columns:} From left to right, ResNet on Waterbirds, ResNet on CelebA Blond / Male, ResNet on CelebA Lipstick / Earring, BERT on MultiNLI.}
\label{fig:summary}
\end{figure}

We are interested in understanding how worst-group test accuracy changes with model size. 
To investigate the trends systematically, we study the effect of varying width and depth, under pre-trained and randomly initialized models. We perform four sets of experiments, varying the depth and width of pre-trained and randomly initialized models. The summarized results of the experiments on the Waterbirds and MultiNLI datasets are presented in Figure~\ref{fig:summary}.

\subsection{Analysis}
\textbf{Pre-trained Models and Randomly Initialized Models, Varying Last Layer Width.}
We replicate experiments in prior work and extend results with an additional dataset to show that our setup is comparable. This set of experiments follows the setup of \citet{Sagawa2020AnIO}.
Performance on the Waterbirds and CelebA Blond / Male datasets gives us results consistent with prior work: worst-group error dips close to 50\% for larger widths. Furthermore, we also study the CelebA Lipstick / Earring dataset and the worst-group error approaches 30\% as we increase the model width. The results are summarized in Figure \ref{fig:reproduce}. Overall, we find that worst-group error decreases slightly for both CelebA datasets before saturating. On Waterbirds, the worst-group error decreases as we increase the number of features and remains around 50\% when the number of features is larger than $10^3$.

\textbf{Pre-trained Models, Varying Depth and Width.} For pre-trained models, we find that model size does not hurt worst-group test error over multiple model architecture series. We compare model sizes by varying depth on ResNet, VGG BN, and BERT; and width on ResNet18, MobileNet, and BERT. On the Waterbirds and MultiNLI datasets, larger models monotonically improve worst-group accuracy. For the two CelebA datasets, increasing model sizes neither significantly increases nor decreases the worst-group test error. Results for varying depth of ResNet and VGG BN models are summarized in Figure~\ref{fig:pretrain-depth}; Resnet18 with varying width is given in Figure~\ref{fig:pretrain-width}; and BERT results are given in Figure~\ref{fig:bert-depth-width}.

\textbf{Randomly Initialized Models, Varying Depth and Width.} For models trained from scratch, we also find that model size does not hurt worst-group test error. However, unlike the pre-trained results, the error did not improve with the increasing model size. %
ResNet results can be seen in Figure \ref{fig:scratch-depth} for varying depth and Figure \ref{fig:scratch-width} for varying width, while BERT results can be seen in Figure \ref{fig:bert-depth-width}.

\subsection{Experimental Takeaways}
Empirically, worst-group test error decreases or stays the same as model size increases. 
The experiments show that pre-trained models perform significantly better than those trained from scratch. 
On the Waterbirds and MultiNLI datasets we find that increasing the size of the pre-trained model improves performance on the worst group. This suggests that pretraining may be a factor in improving performance of overparameterized models.

On the other hand, we show in Appendix \ref{table:dro} that the worst-group performance of group DRO (which uses group label information) is generally better than that of ERM. Therefore, group DRO should be used when group labels are available, and our analysis primarily applies to the case where they are unavailable.

%% file: parts/5_Conclusion.tex
\section{Conclusion}\label{conclusion}

Although increasing model size only sometimes helps worst-group generalization, large models generally do not hurt across almost all the ERM settings, whether the model is pre-trained or trained from scratch. %
Furthermore, we find that as compared to models trained from scratch, increasing pre-trained model size is more likely to improve worst-group accuracy. 
We leave for future work the effects of pre-training on the worst-group performance.
Our experimental results suggest that the study of spurious correlations under the ERM setting is interesting from both the practical and analytical perspectives, and it can potentially lead to novel ways of designing experimental protocols and algorithms.

\section*{Acknowledgments}

We would like to thank Pang Wei Koh, Aditi Raghunathan, and Erik Jones for their valuable feedback and comments during preparation of this manuscript. Joseph E. Gonzalez would like to acknowledge supports from NSF CISE Expeditions Award CCF-1730628 and gifts from Alibaba, Amazon Web Services, Ant Group, Ericsson, Facebook, Futurewei, Google, Intel, Microsoft, Nvidia, Scotiabank, Splunk and VMware.
Our conclusions do not necessarily reflect the position or the policy of our sponsors, and no official endorsement should be inferred.

%% file: parts/7_Appendix.tex
\begin{figure}[ht]
\begin{multicols}{2}[\columnsep=2.5cm] %
\includegraphics[scale=0.21]{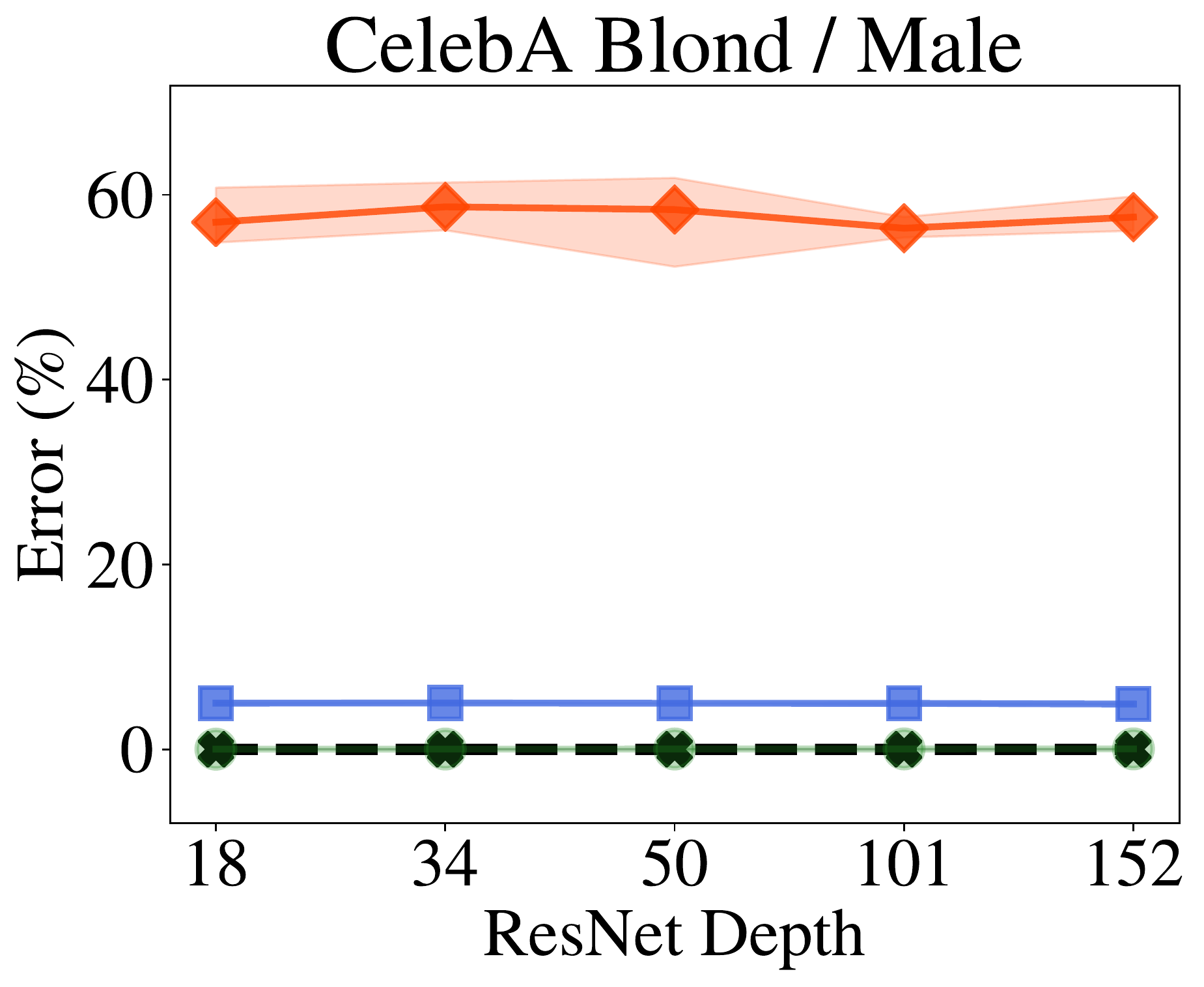}
\columnbreak
\includegraphics[scale=0.21]{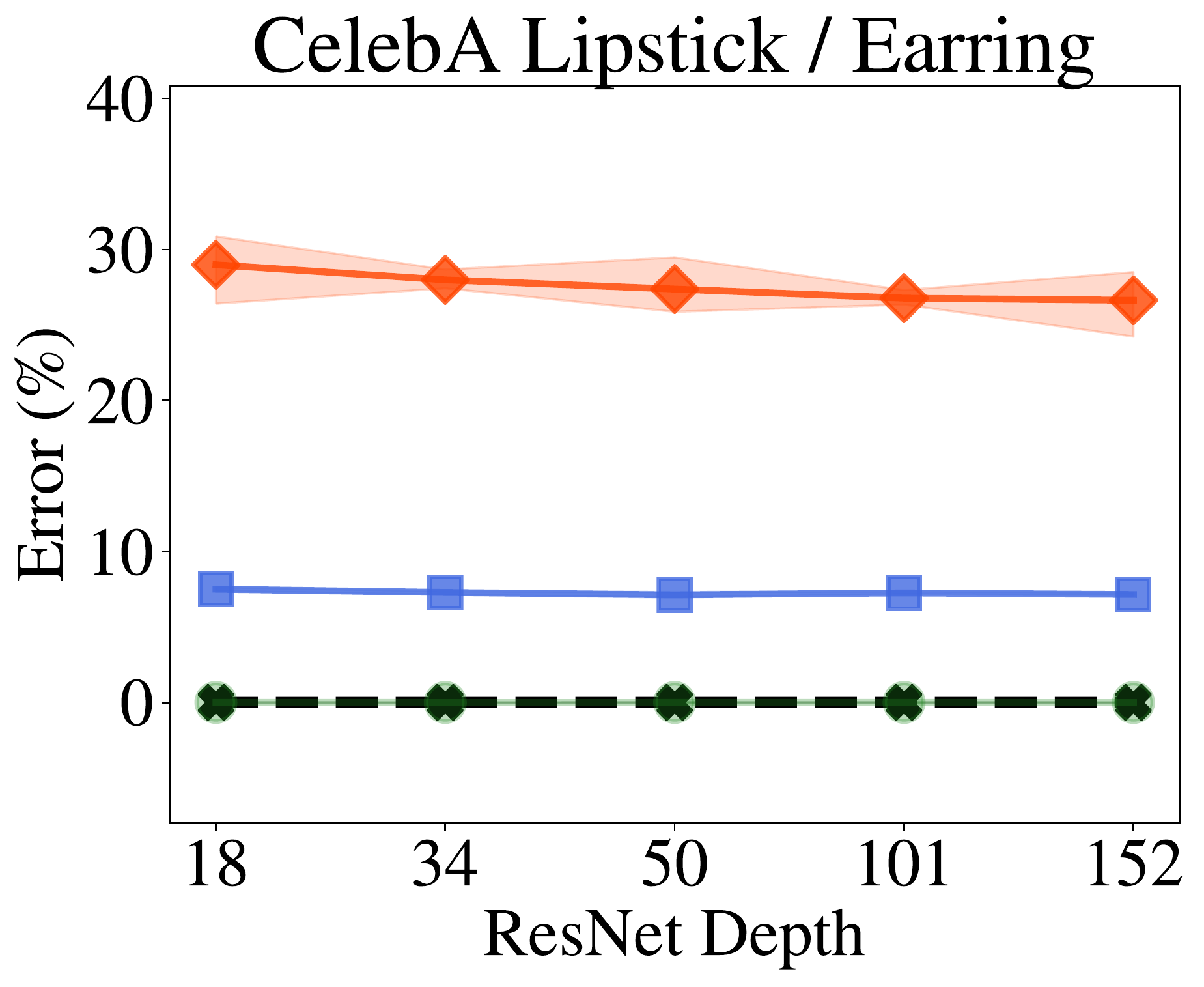}
\columnbreak
\includegraphics[scale=0.21]{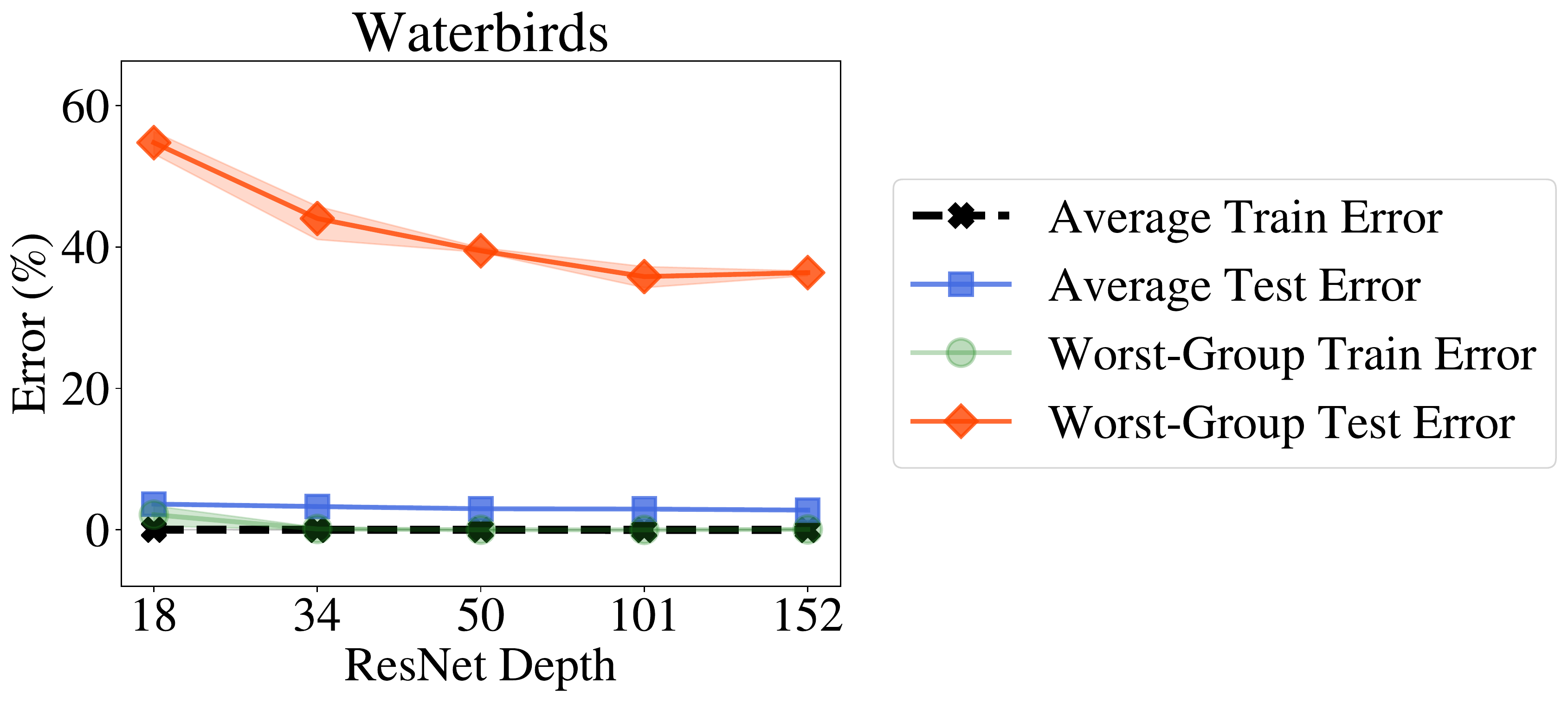}
\end{multicols}

\begin{multicols}{2}[\columnsep=2.5cm] %
\includegraphics[scale=0.21]{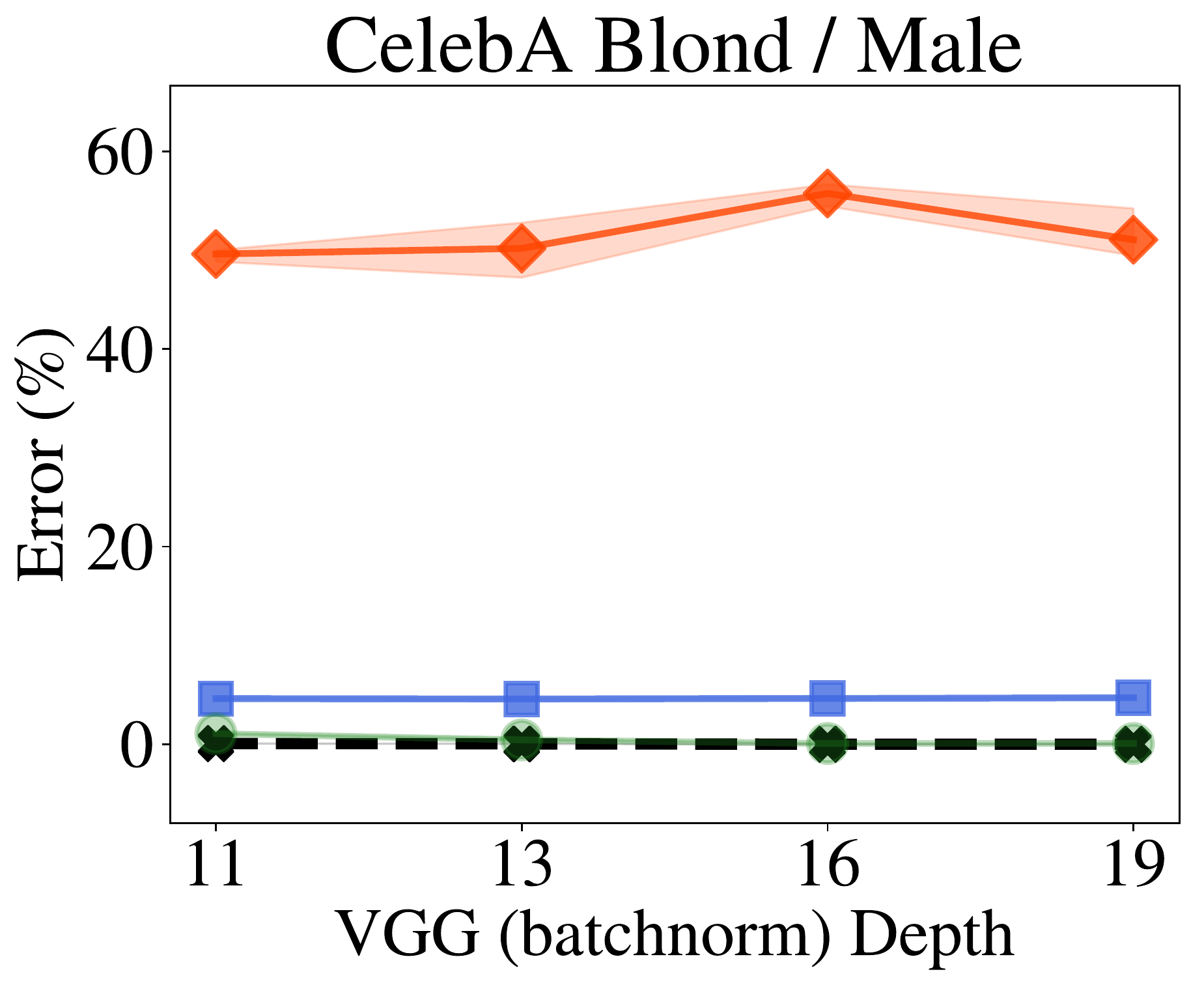}
\columnbreak
\includegraphics[scale=0.21]{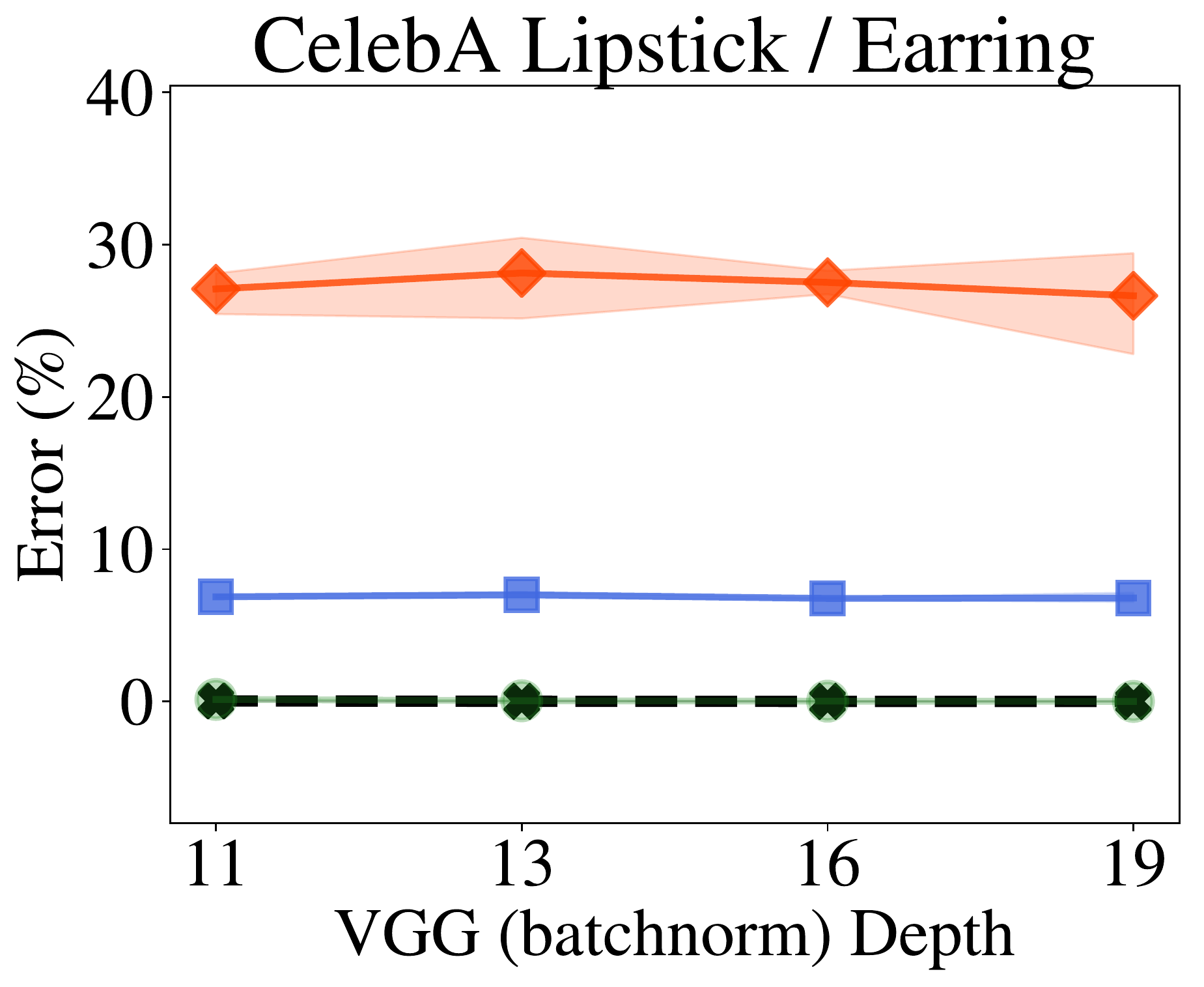}
\columnbreak
\includegraphics[scale=0.21]{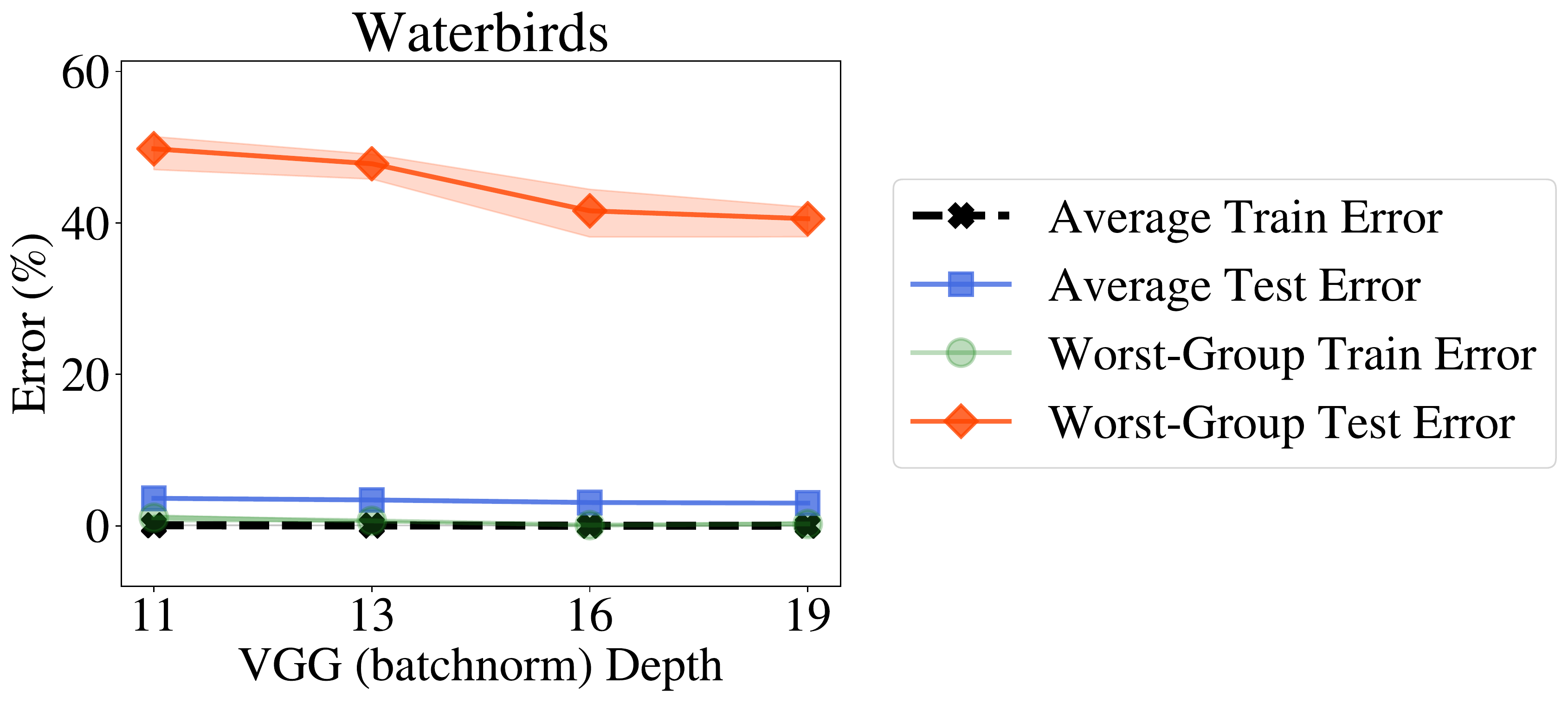}
\end{multicols}
\vspace{-0.15in}
\caption{The top row shows the pre-trained ResNet models of varying depth. The bottom row shows the pre-trained VGG models of varying depth. Each column represents the dataset the model is trained and evaluated on. From left to right: CelebA Blond / Male, CelebA Lipstick / Earring, and Waterbirds. For the two CelebA datasets, model depth has a negligible effect on the worst-group error whereas on the Waterbirds dataset, the increasing the model size decreases the worst-group error.}
\label{fig:pretrain-depth}
\end{figure}
\begin{figure}[H]
\begin{multicols}{2}[\columnsep=2.5cm] %
\includegraphics[scale=0.21]{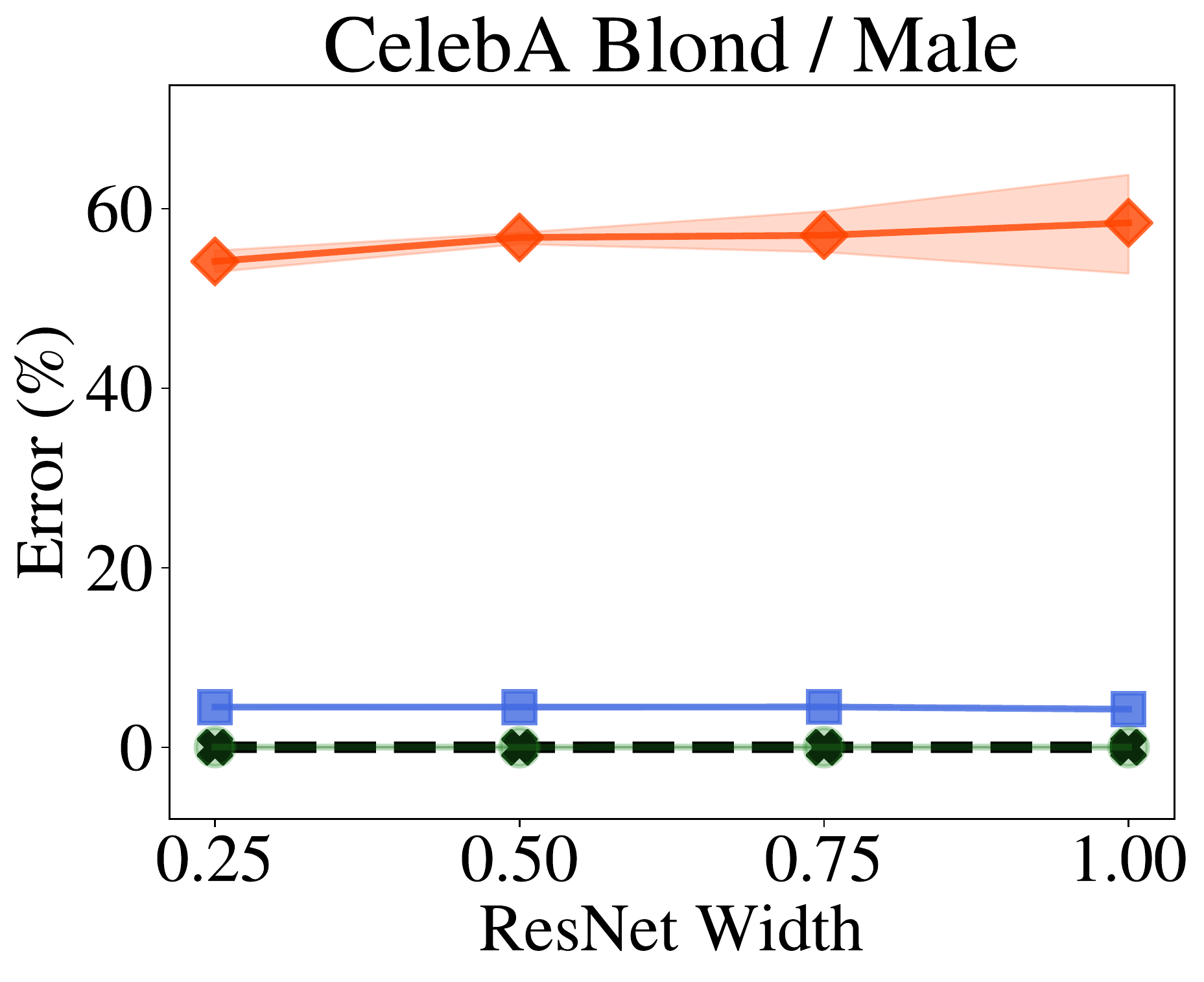}
\columnbreak
\includegraphics[scale=0.21]{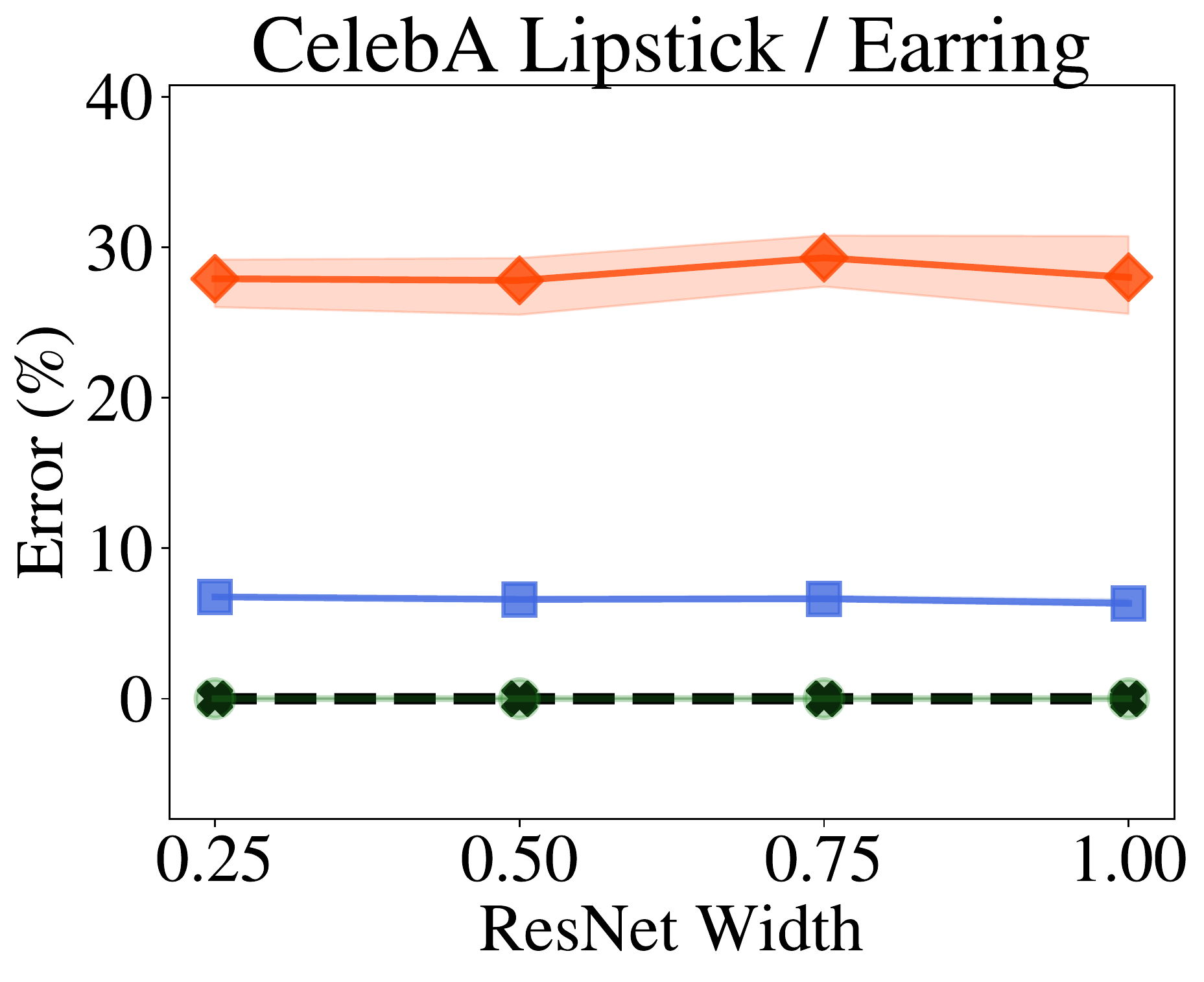}
\columnbreak
\includegraphics[scale=0.21]{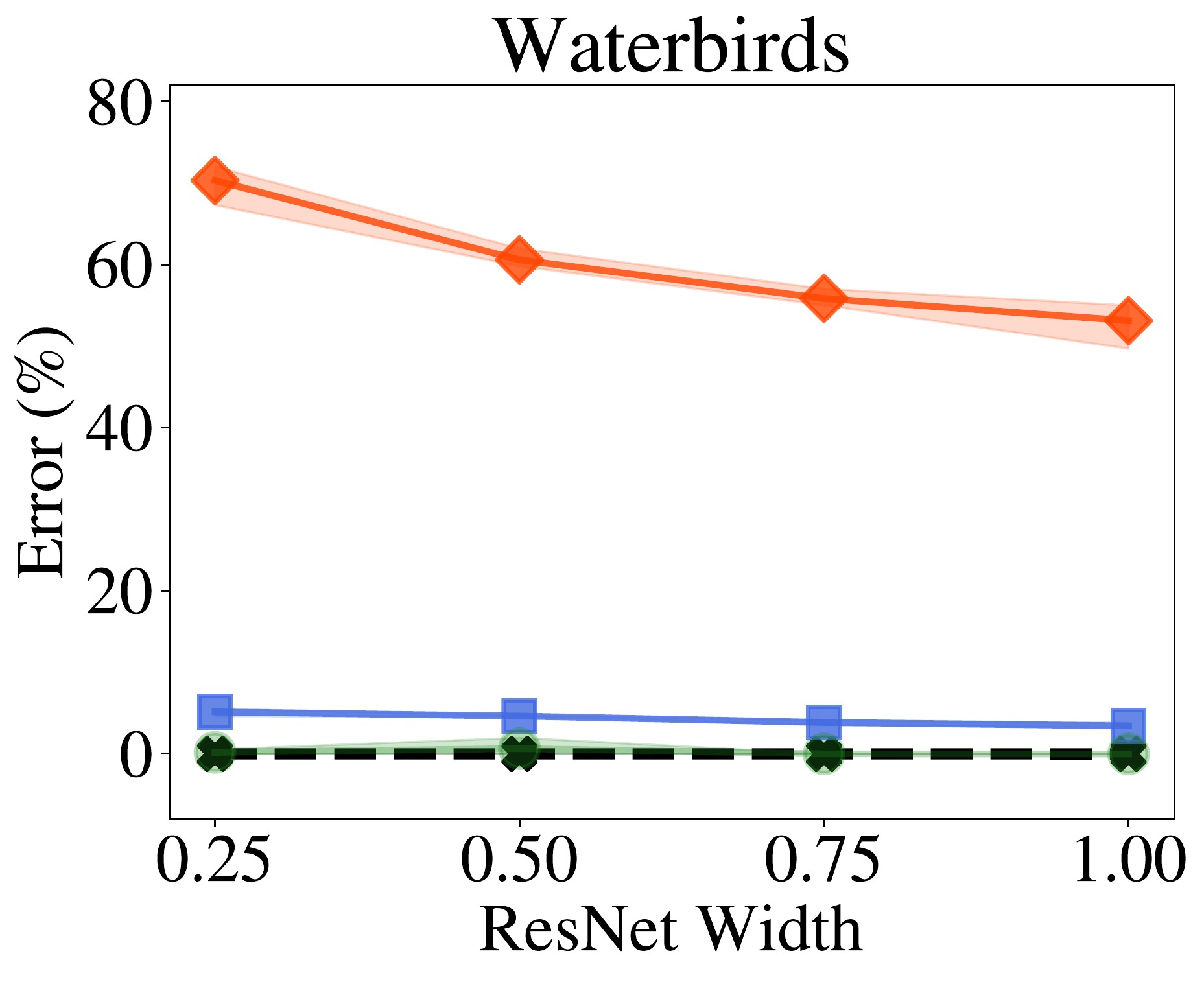}
\end{multicols}

\begin{multicols}{2}[\columnsep=2.5cm] %
\includegraphics[scale=0.21]{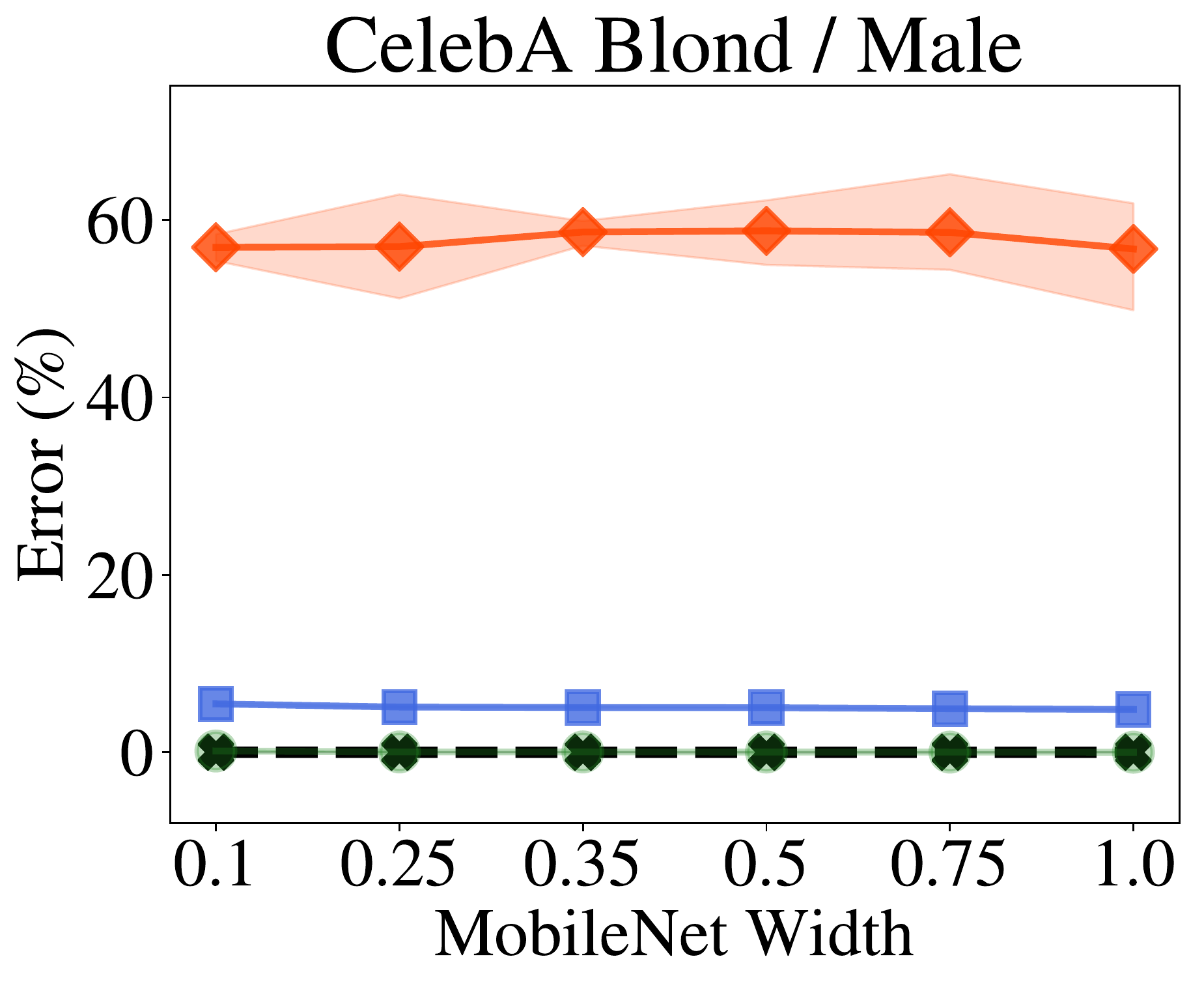}
\columnbreak
\includegraphics[scale=0.21]{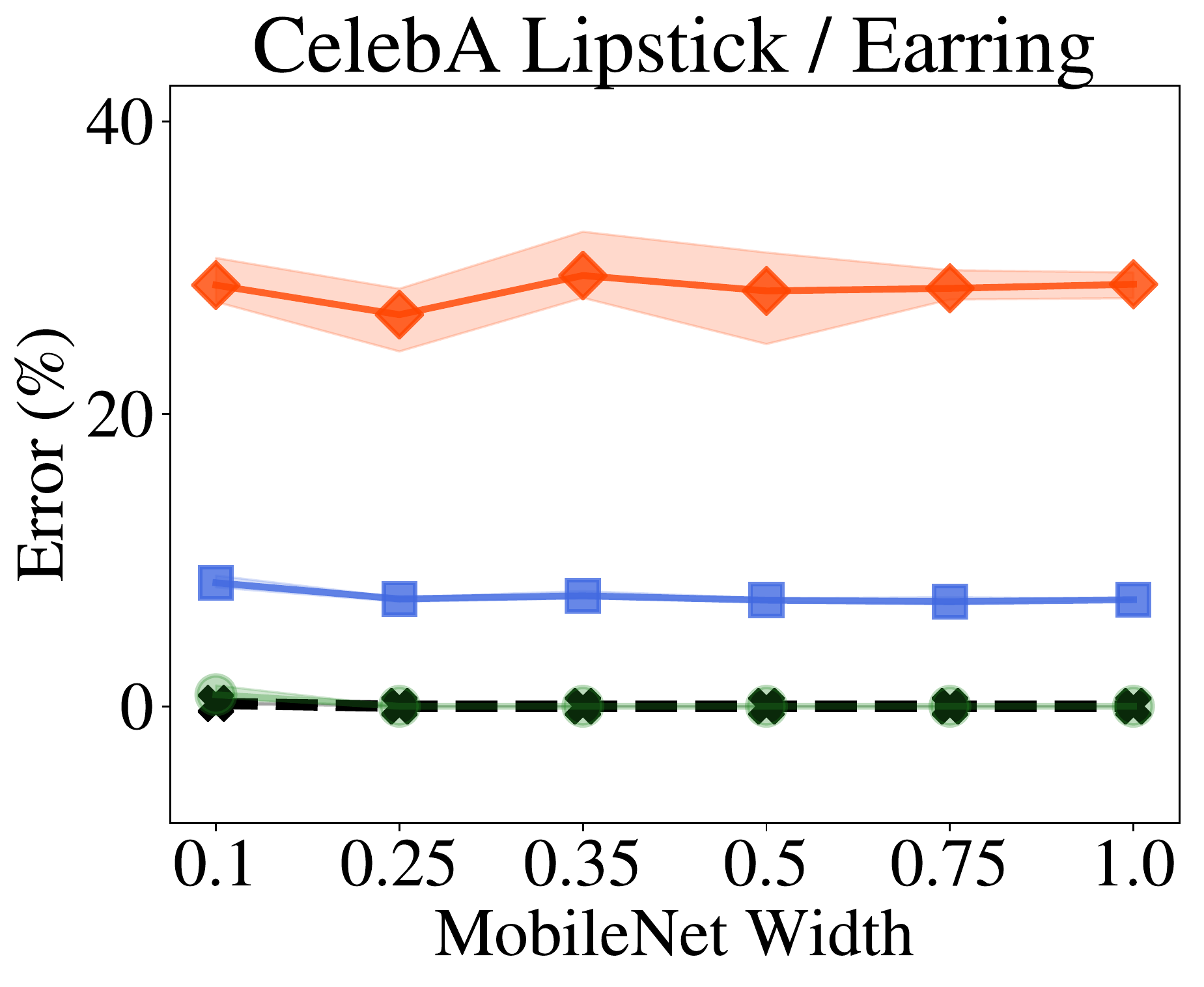}
\columnbreak
\includegraphics[scale=0.21]{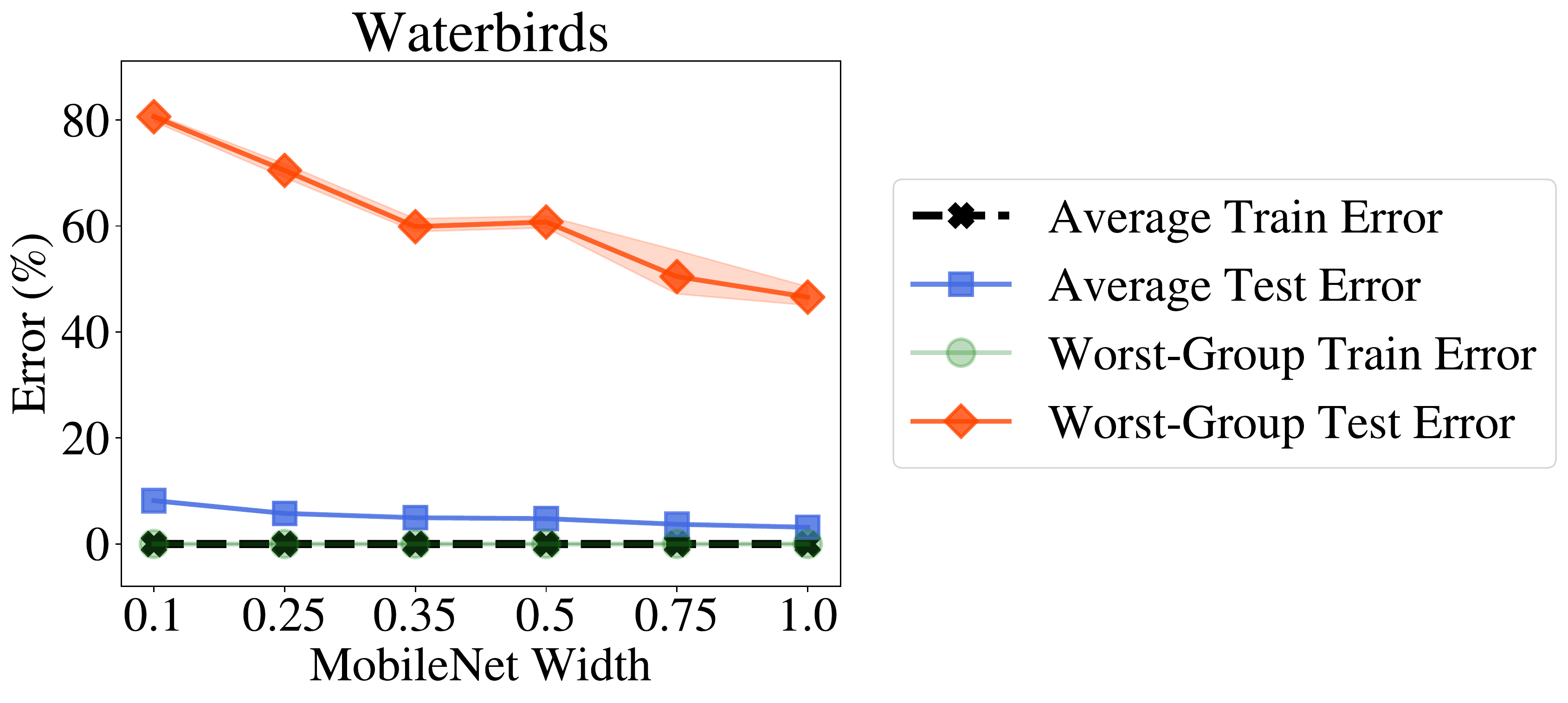}
\end{multicols}
\vspace{-0.15in}
\caption{The top row shows the pre-trained ResNet18 models of varying widths. The bottom row shows the pre-trained MobileNet models of varying width. Each column represents the dataset the model is trained and evaluated on. From left to right: CelebA Blond / Male, CelebA Lipstick / Earring, and Waterbirds. For the two CelebA datasets, model depth has a negligible effect on the worst-group error whereas on the Waterbirds dataset, the increasing the model size decreases the worst-group error.} 
\label{fig:pretrain-width}
\vspace{-0.1in}
\end{figure}
\begin{figure}
\begin{multicols}{2}[\columnsep=2.5cm] %
\includegraphics[scale=0.21]{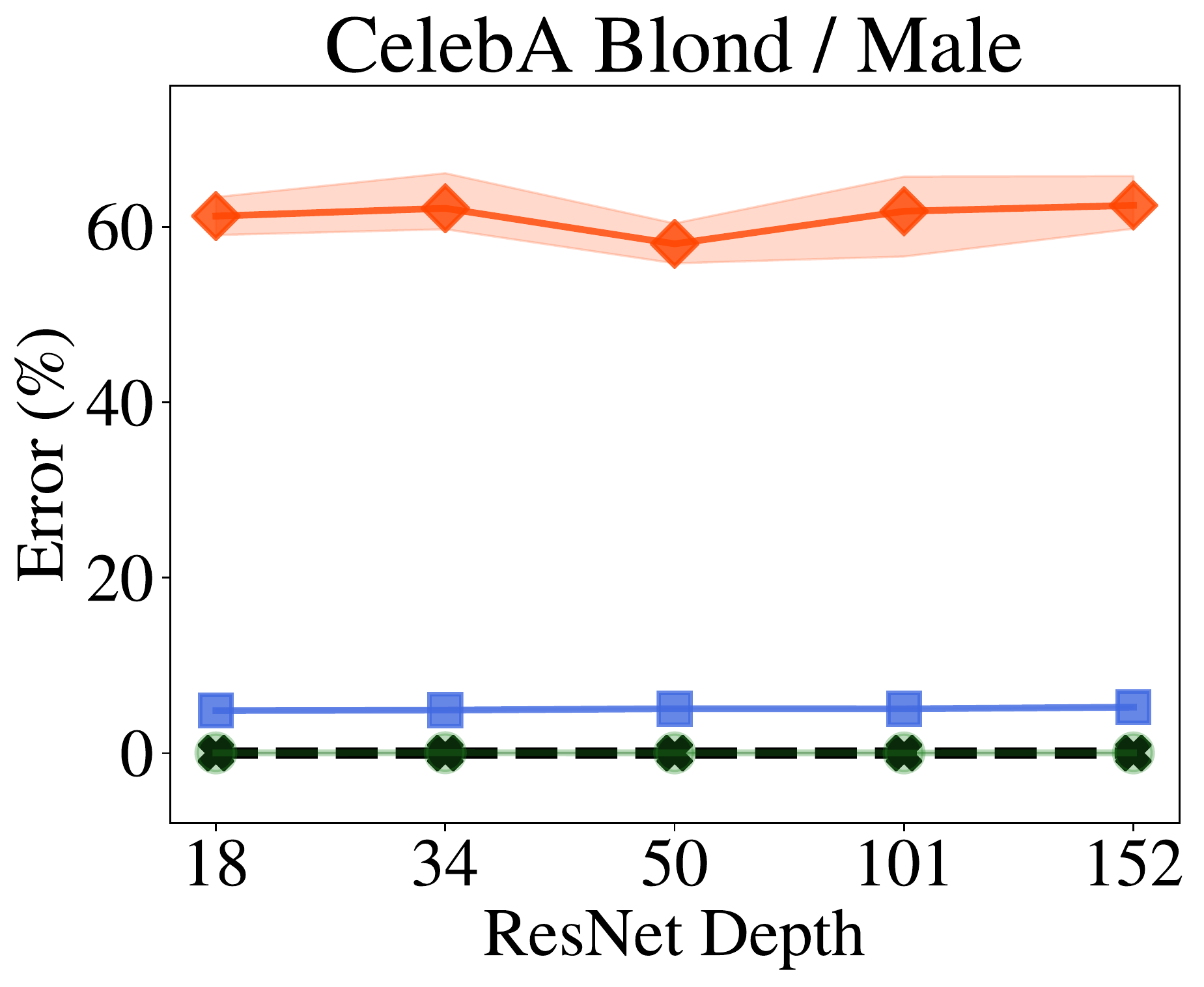}
\columnbreak
\includegraphics[scale=0.21]{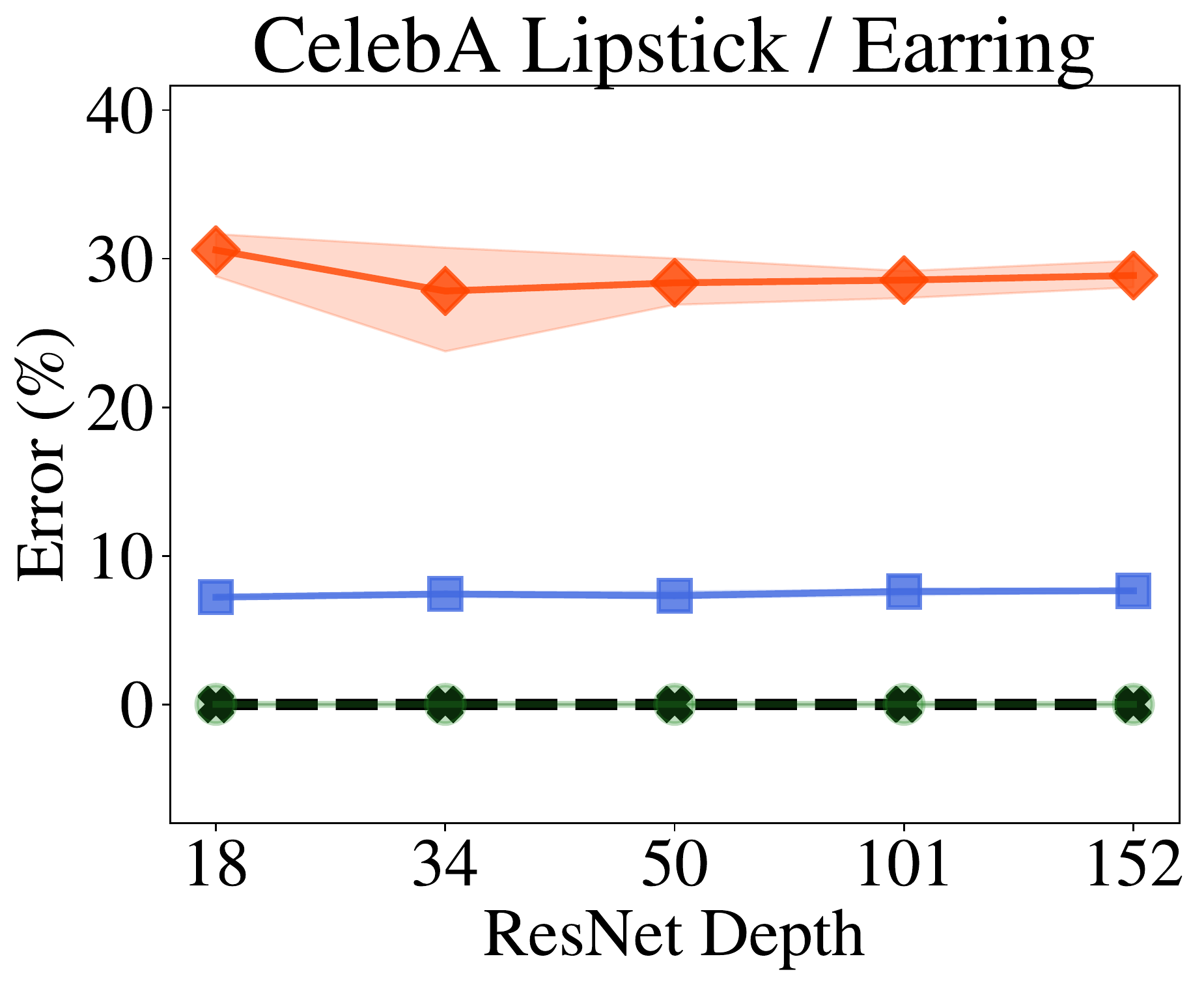}
\columnbreak
\includegraphics[scale=0.21]{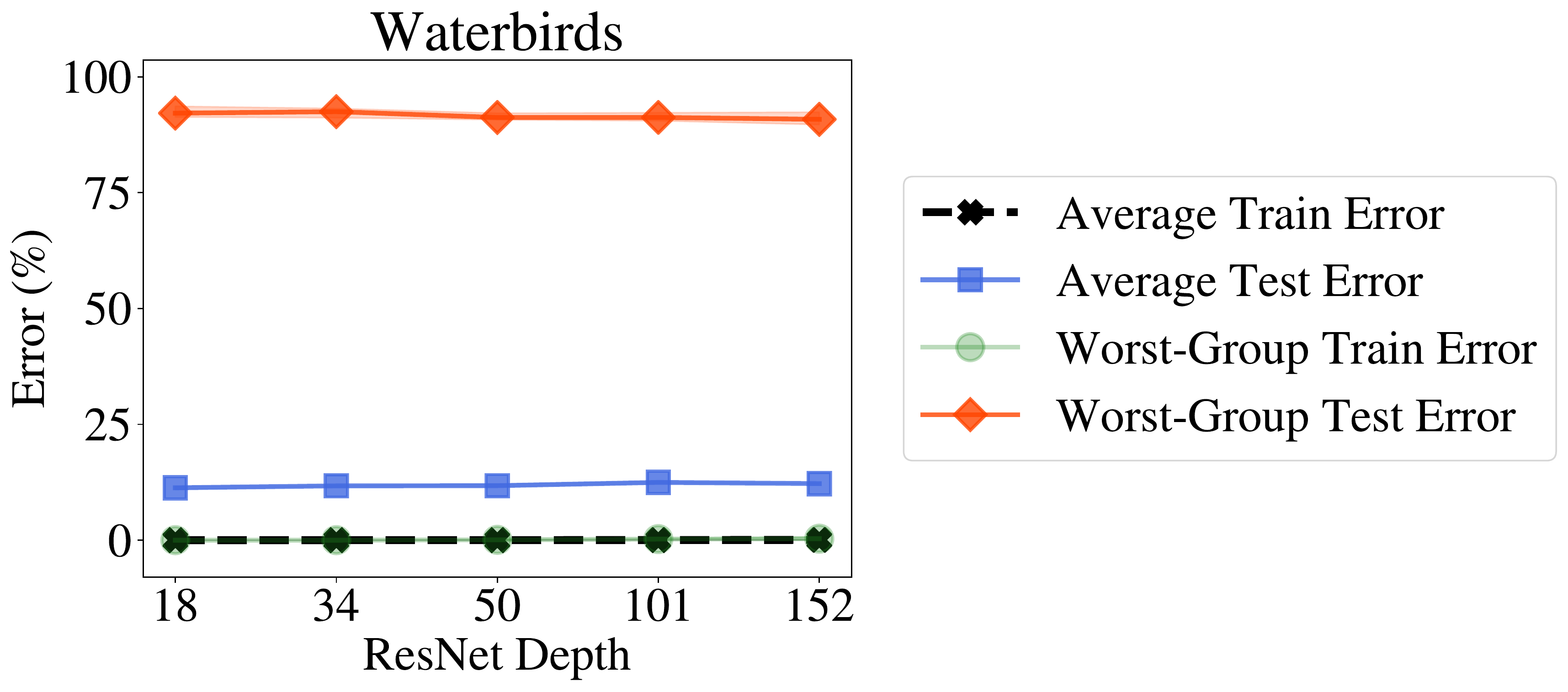}
\end{multicols}
\caption{Depth of randomly initialized ResNet models is varied, increasing in depth from left to right. Each column represents the dataset the model is trained and evaluated on. From left to right: CelebA Blond / Male, CelebA Lipstick / Earring, and Waterbirds. For all of the datasets, model depth has a negligible effect on the worst-group error.}
\label{fig:scratch-depth}
\end{figure}
\begin{figure}
\begin{multicols}{2}[\columnsep=2.5cm] %
\includegraphics[scale=0.21]{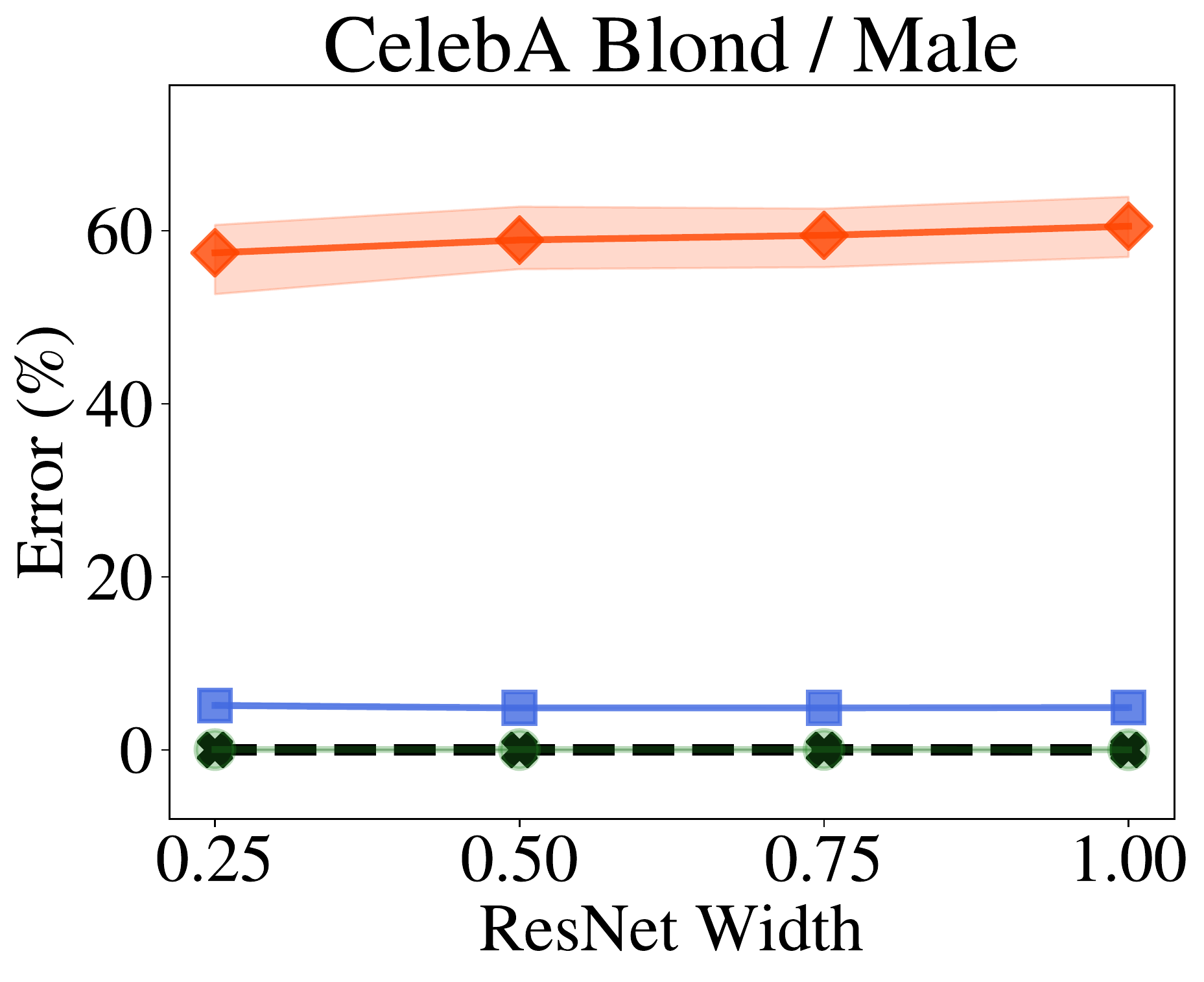}
\columnbreak
\includegraphics[scale=0.21]{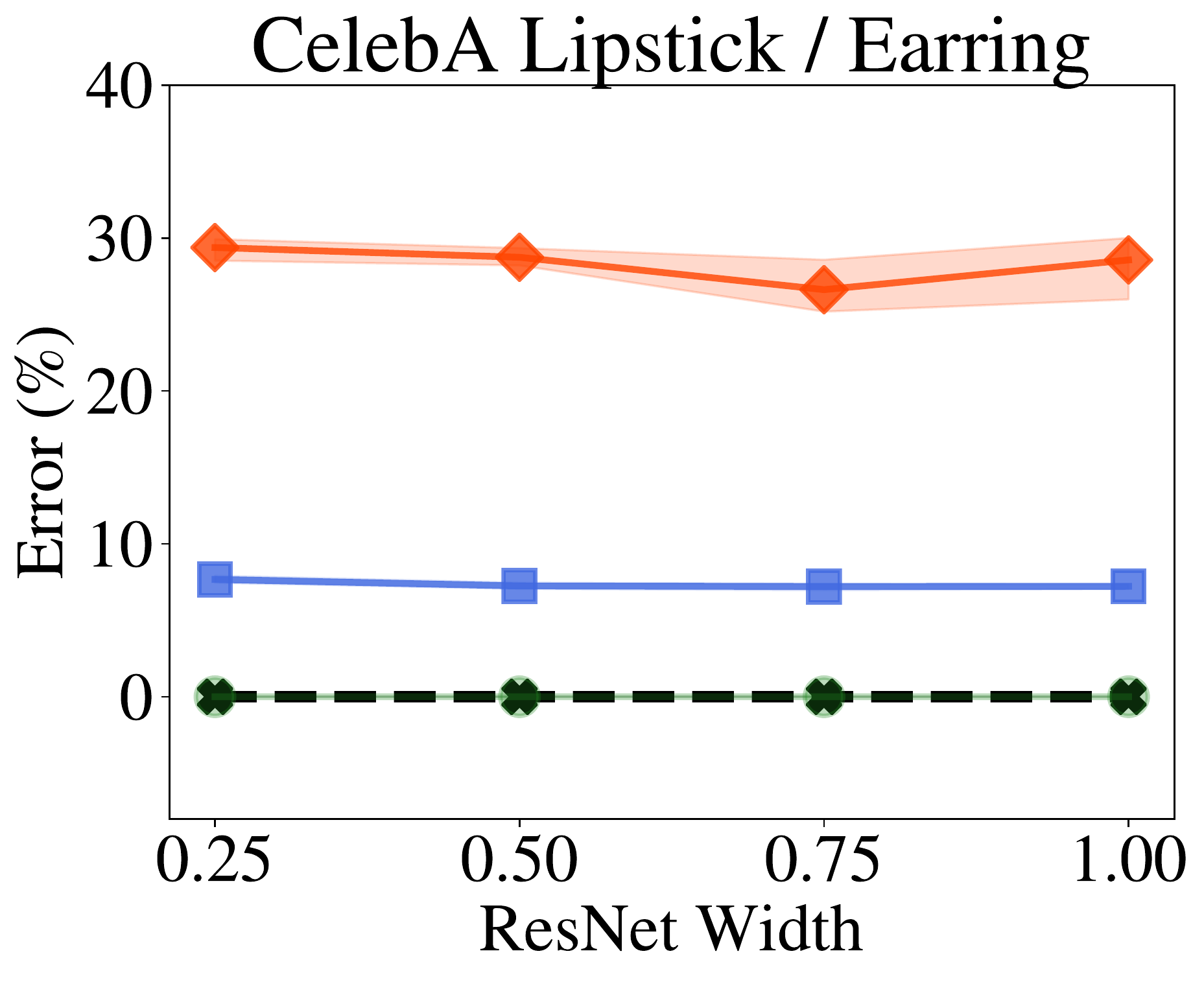}
\columnbreak
\includegraphics[scale=0.21]{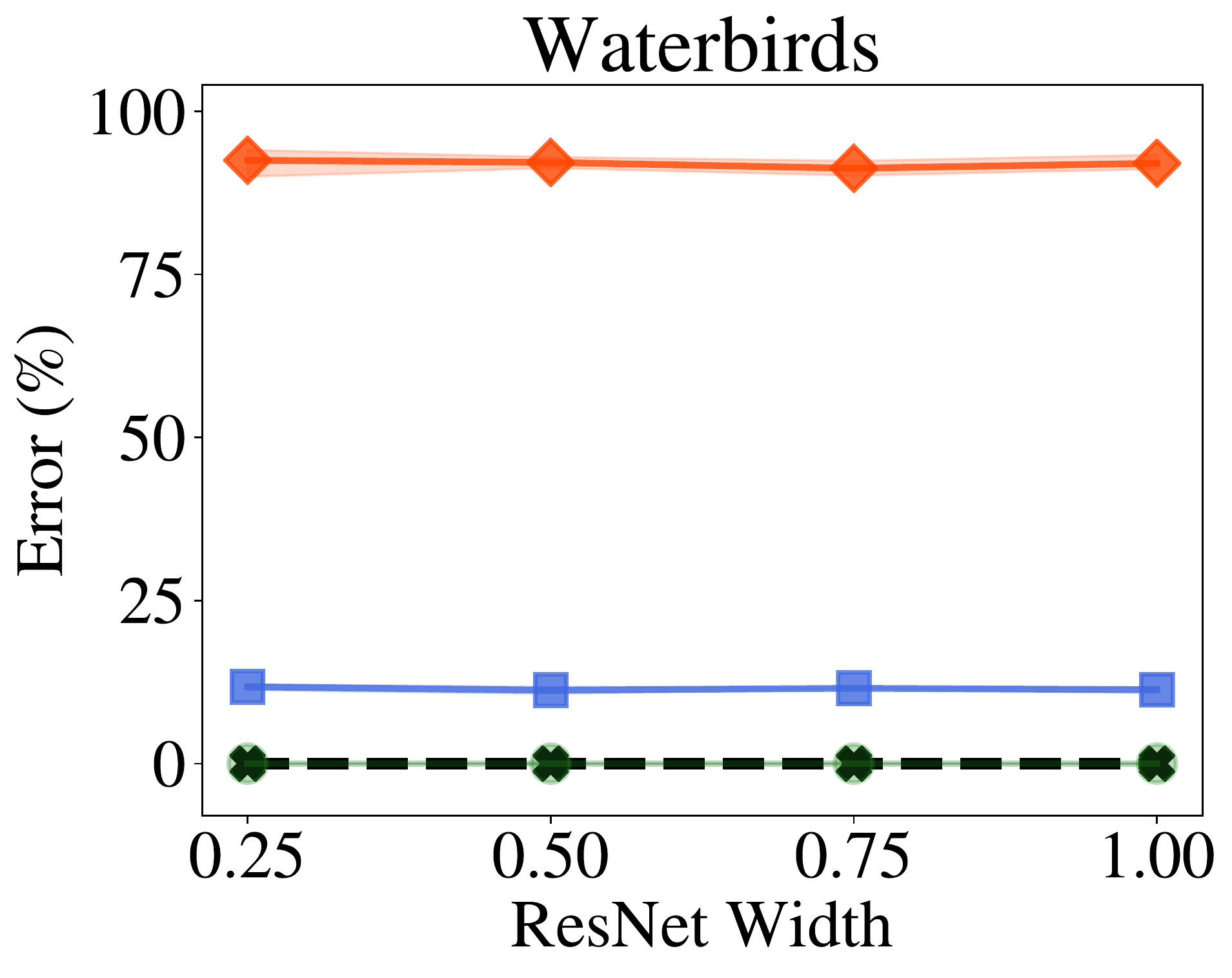}
\end{multicols}

\caption{Width of randomly initialized ResNet models is varied, increasing in width from left to right. Each column represents the dataset the model is trained and evaluated on. From left to right: CelebA Blond / Male, CelebA Lipstick / Earring, and Waterbirds. For all of the datasets, model depth has a negligible effect on the worst-group error.}
\label{fig:scratch-width}
\end{figure}

We summarize our depth varying for pre-trained models results in figure \ref{fig:pretrain-depth}, width varying for pre-trained models in figure \ref{fig:pretrain-width}, depth varying for randomly initialized models in figure \ref{fig:scratch-depth}, and width varying for randomly initialized models in figure \ref{fig:scratch-width}. Pre-trained models on the Waterbirds dataset show a distinct decrease in error with model size while for other experiments, error stays roughly the same for all model sizes. 


\begin{figure}
\begin{multicols}{2}[\columnsep=2.5cm] %
\includegraphics[scale=0.21]{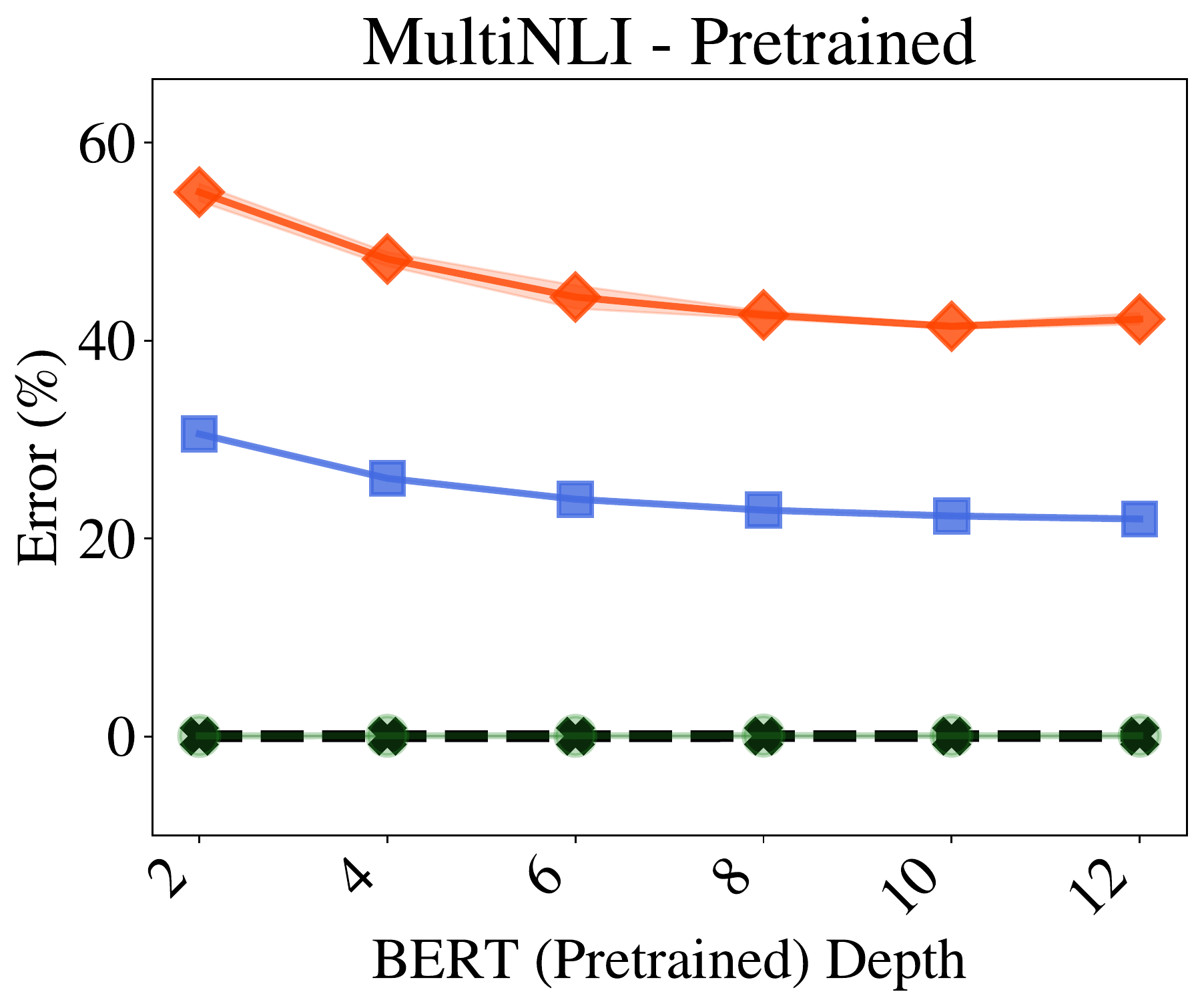}
\columnbreak
\includegraphics[scale=0.21]{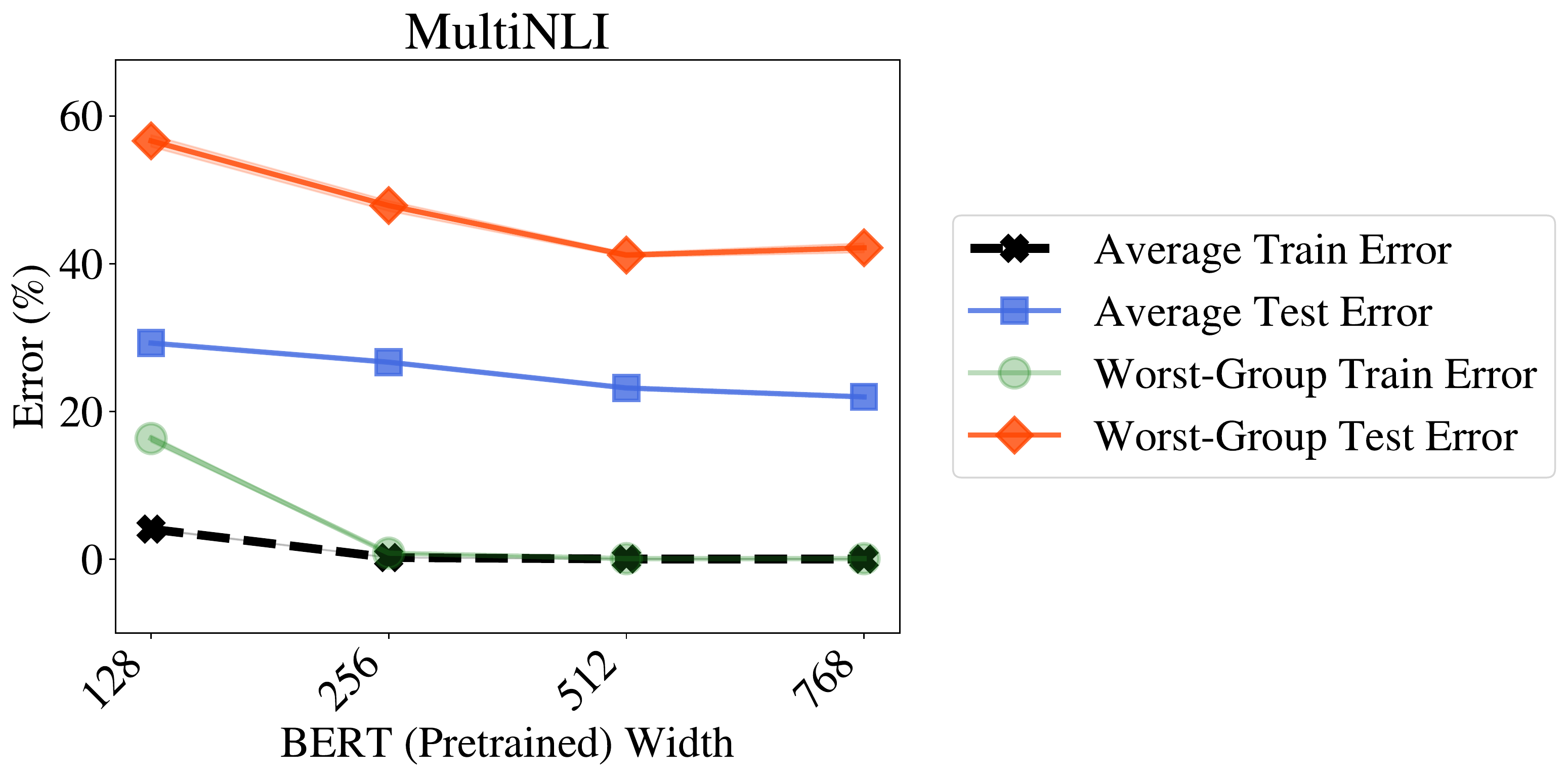}
\end{multicols}
\begin{multicols}{2}[\columnsep=2.5cm] %
\includegraphics[scale=0.21]{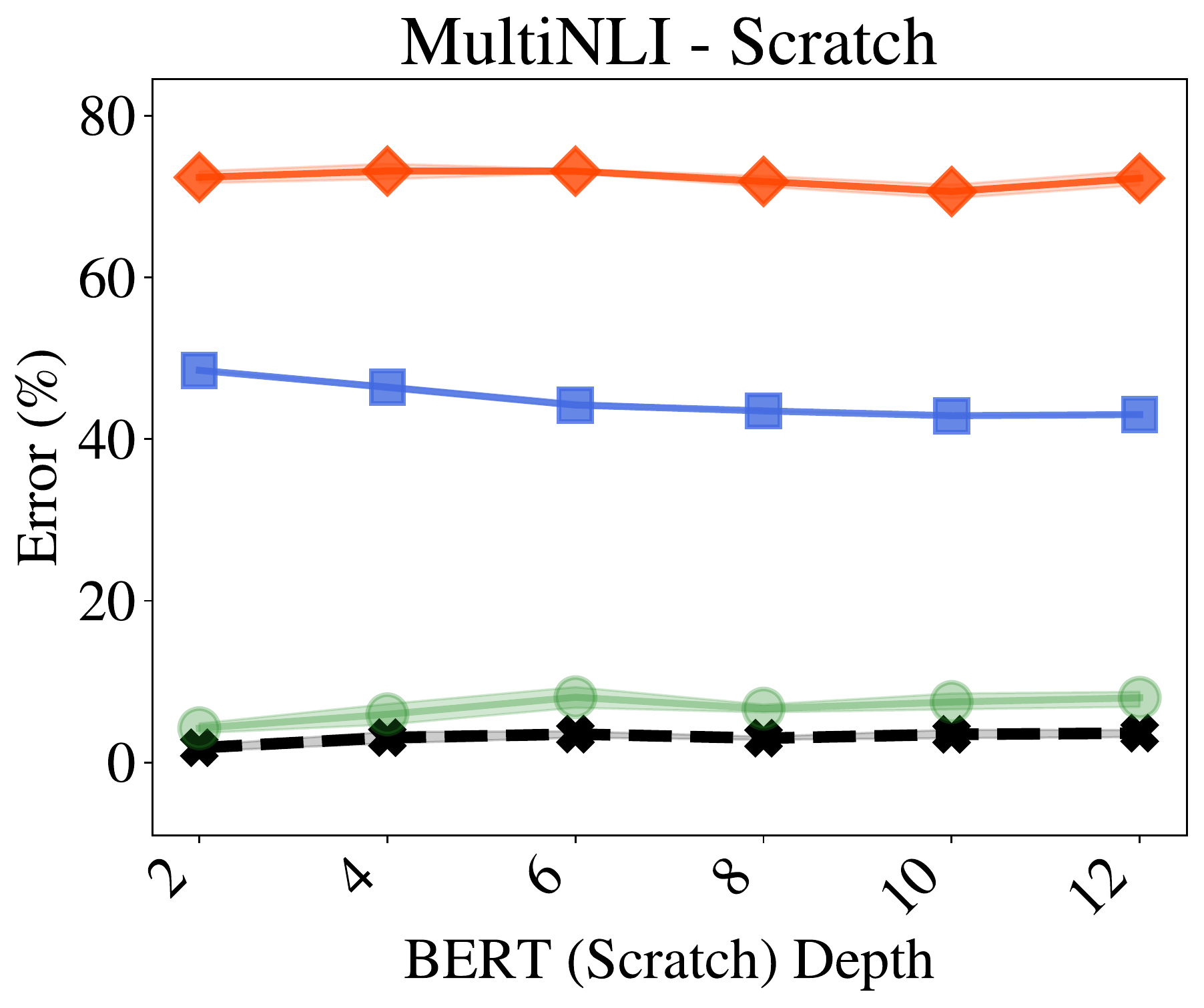}
\columnbreak
\includegraphics[scale=0.21]{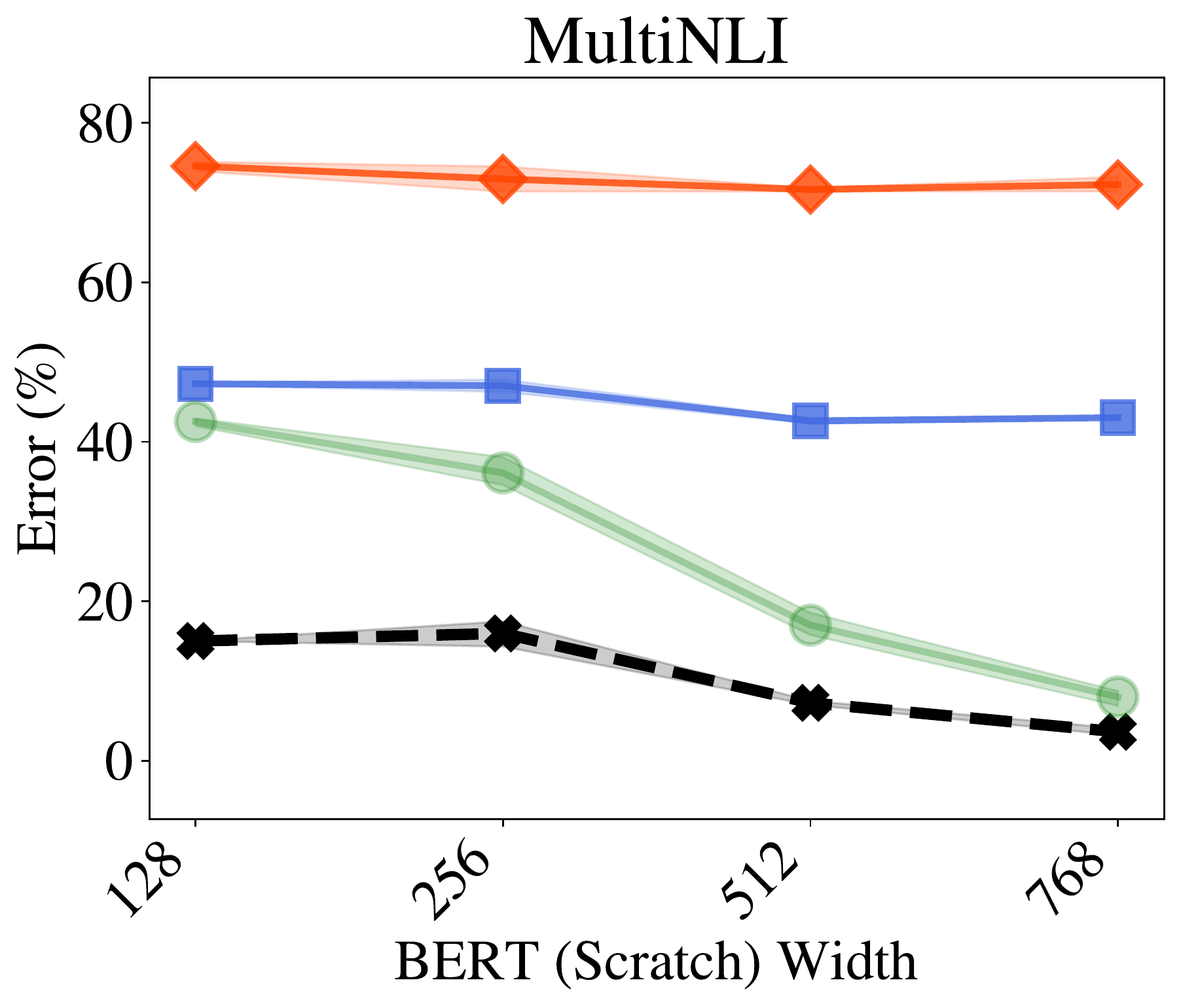}
\end{multicols}

\caption{\textbf{Top Row:} Depth and width of pre-trained BERT models are varied, increasing in size from left to right. \textbf{Bottom Row:} Depth and width of randomly initialized BERT models are varied, increasing in size from left to right. Increasing pre-trained model size reduces worst-group error, while on randomly initialized models, model size has negligible effect.}
\label{fig:bert-depth-width}
\end{figure}

\begin{figure}
\begin{multicols}{1}[\columnsep=2.5cm] %
\includegraphics[scale=0.21]{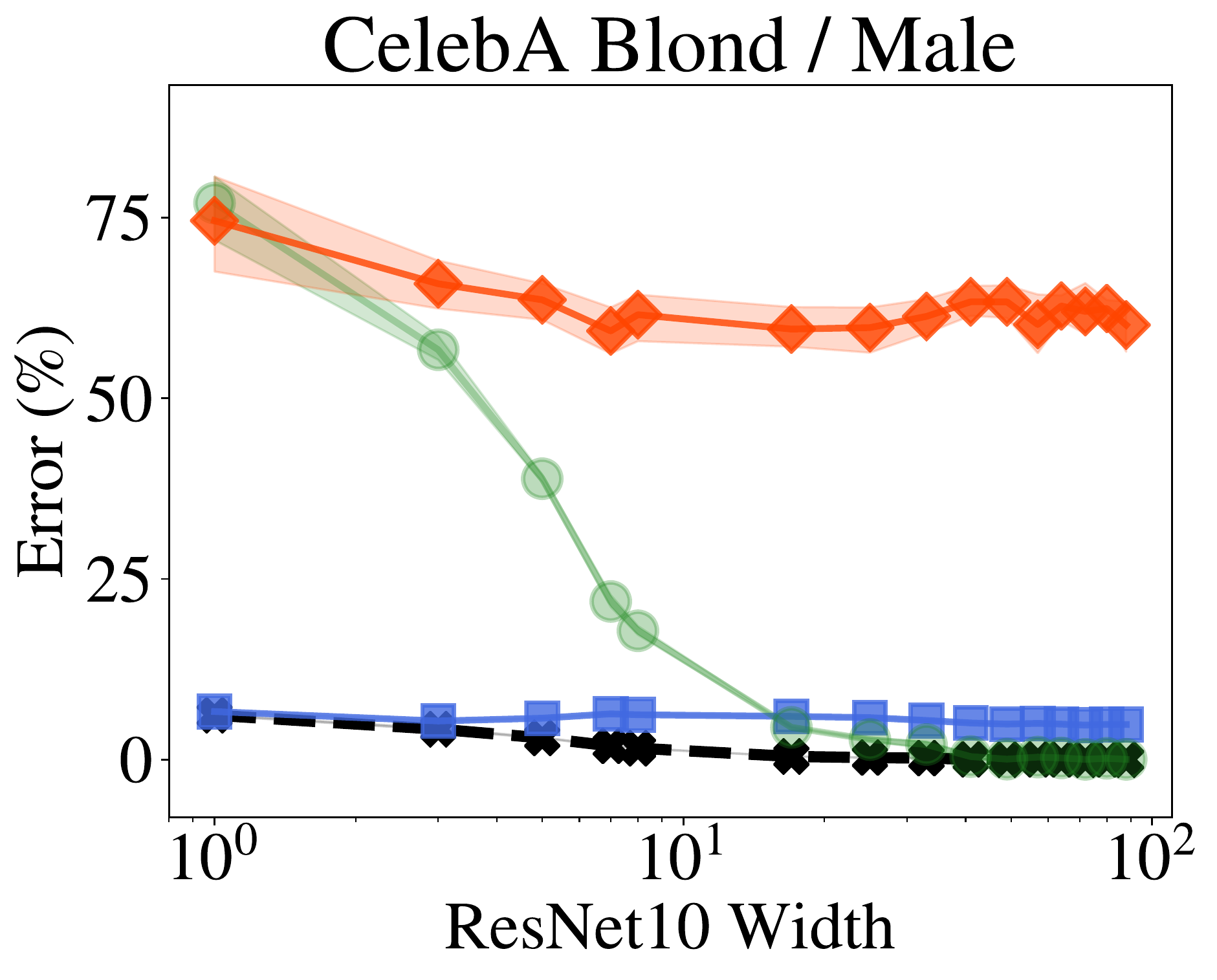}
\columnbreak
\includegraphics[scale=0.21]{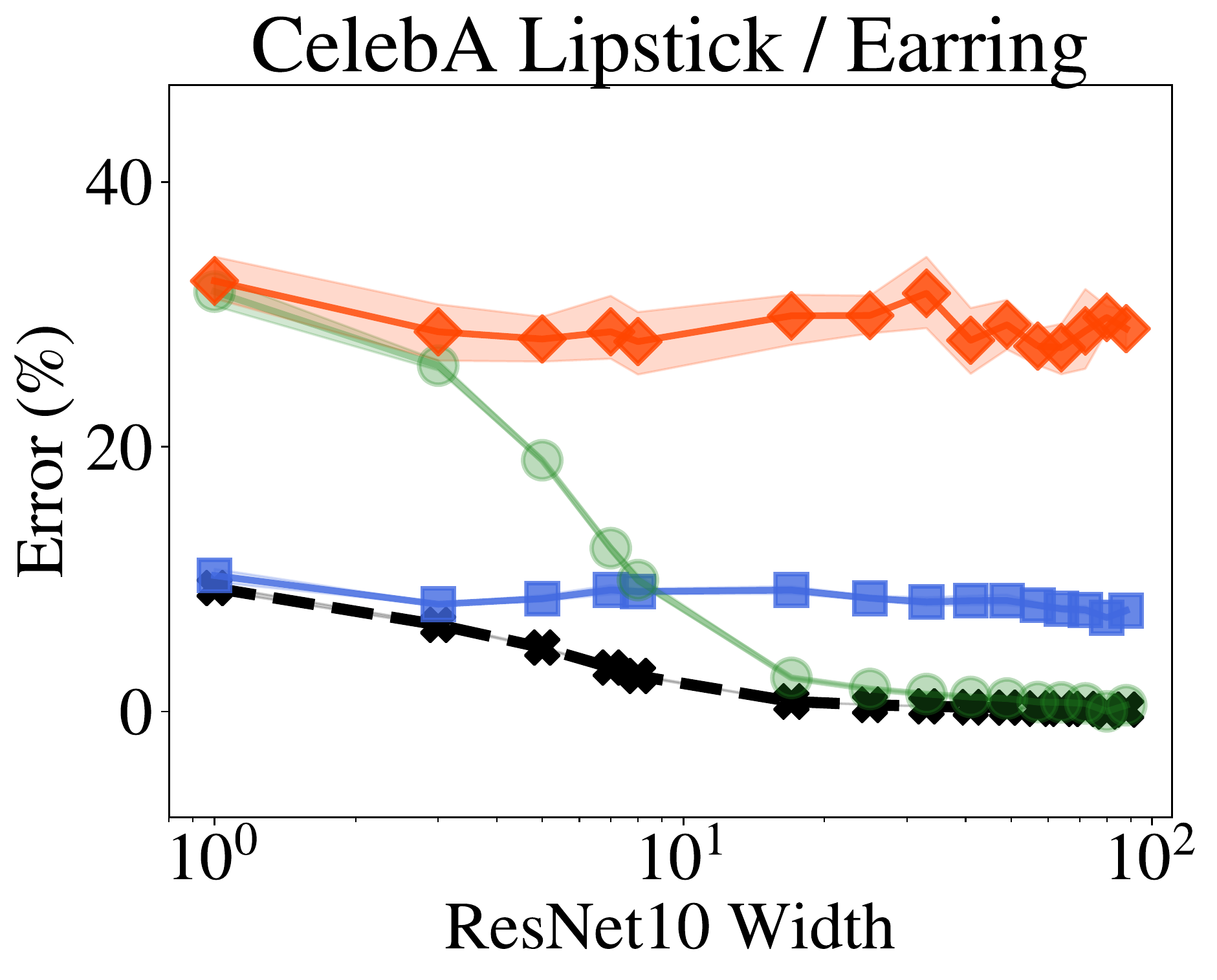}
\columnbreak
\includegraphics[scale=0.21]{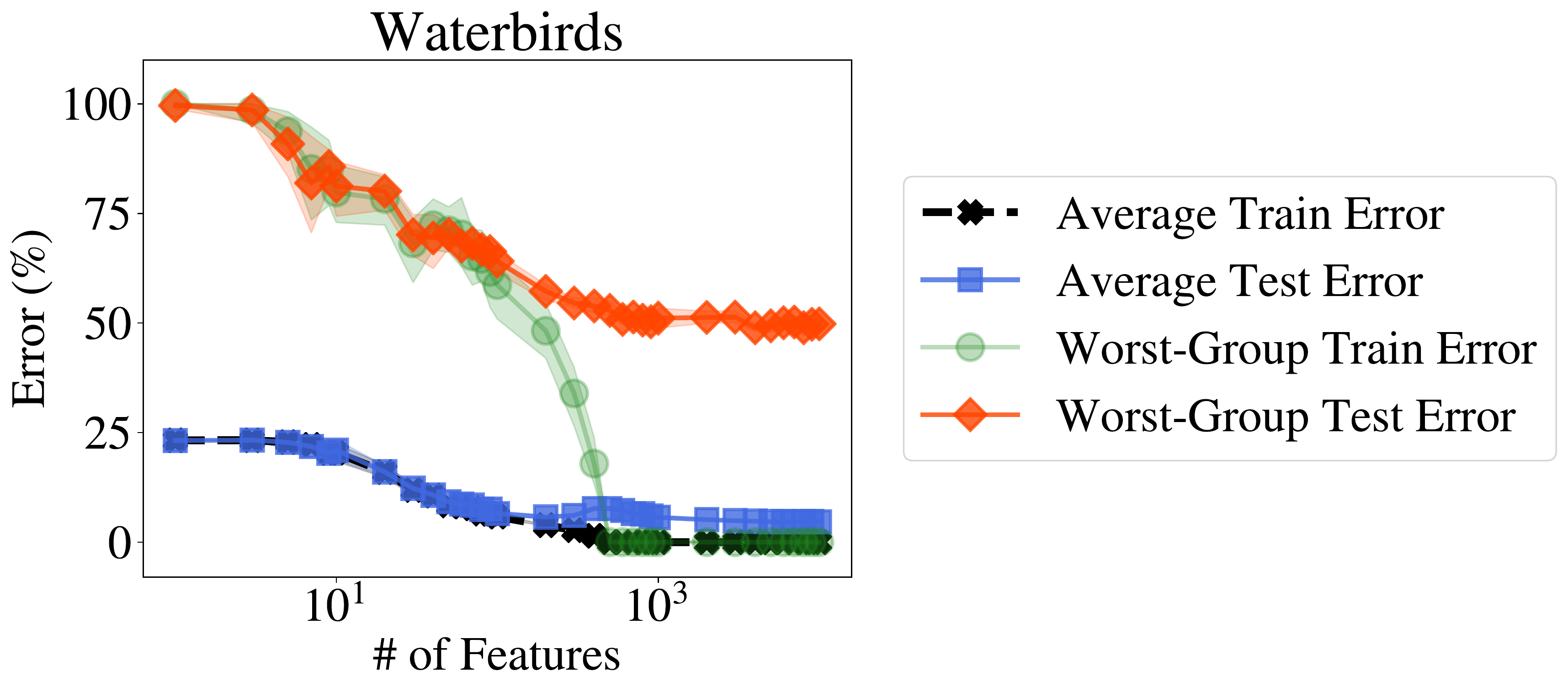}
\end{multicols}
\caption{Graphs displaying how the error changes as we vary the width or number of features. From left to right: CelebA Blond / Male, CelebA Lipstick / Earring, and Waterbirds. For CelebA Blond / Male and Waterbirds, worst-group error improves with the greater model size whereas worst-group error on the CelebA Lipstick / Earring seems to stay at about the same value as model size increases.}
\label{fig:reproduce}
\end{figure}

\newpage
\input{parts/6_Framework}

\begin{table}[h!]
\centering
\label{table:counts}
\begin{tabular}{| c | c | c | c |}
 \hline
 \textbf{Dataset} & \textbf{Worst-Group} & \textbf{Total Count} & \textbf{Worst-Group Count}\\
 \hline
Waterbirds & Waterbird, Land Background & 4795 & 56 \\ 
\hline
CelebA Blond / Male & Blond, Male & 162770 & 1387 \\ 
\hline
CelebA Lipstick / Earring & No Lipstick, Earring & 162770 & 4516 \\ 
\hline
MultiNLI & Entailment, Negation  & 205357 & 1483 \\ 
 \hline
 \end{tabular}
\caption{Number of training examples in each dataset}
\end{table}

\begin{table}[t]
\label{hyperparameters_reddit}
\centering
\begin{tabular}{| c | c | c | c | c | c | c |}
 \hline
 \textbf{Dataset} & \textbf{Inital Weight} & \textbf{Architecture} & \textbf{Epochs} & \textbf{LR} & \textbf{LR Step} & \textbf{WD}\\
 \hline
 \multirow{6}*{Waterbirds} & \multirow{4}*{Pretrained} & ResNet & 100 & 5e-4 & 30 & 1e-4 \\ \cline{3-7}
 && ResNet Width & 100 & 0.001 & 30 & 1e-4 \\ \cline{3-7}
 & & MobileNet & 100 & 0.01 & 30 & 1e-4 \\ \cline{3-7}
 & & VGG BN & 100 & 5e-4 & 30 & 1e-4 \\ \cline{2-7}
 & \multirow{2}*{Random} & ResNet & 100 & 0.01 & 30 & 1e-4 \\ \cline{3-7}
 && ResNet Width & 100 & 0.01 & 30 & 1e-4 \\ 
 \hline
 
  & \multirow{4}*{Pretrained} & ResNet & 100 & 5e-4 & 30 & 1e-4 \\ \cline{3-7}
 && ResNet Width & 100 & 0.001 & 30 & 1e-4 \\ \cline{3-7}
 CelebA & & MobileNet & 200 & 0.05 & 50 & 3e-5 \\ \cline{3-7}
 Blond / Male & & VGG BN & 100 & 5e-4 & 30 & 1e-4 \\ \cline{2-7}
 & \multirow{2}*{Random} & ResNet & 100 & 0.01 & 30 & 1e-4 \\ \cline{3-7}
 && ResNet Width & 100 & 0.01 & 30 & 1e-4 \\ 
 
 \hline
 
 & \multirow{4}*{Pretrained} & ResNet & 100 & 5e-4 & 30 & 1e-4 \\ \cline{3-7}
 && ResNet Width & 100 & 0.001 & 30 & 1e-4 \\ \cline{3-7}
 CelebA & & MobileNet & 200 & 0.06 & 50 & 3e-5 \\ \cline{3-7}
 Lipstick / Earring & & VGG BN & 100 & 5e-4 & 30 & 1e-4 \\ \cline{2-7}
 & \multirow{2}*{Random} & ResNet & 100 & 0.01 & 30 & 1e-4 \\ \cline{3-7}
 && ResNet Width & 100 & 0.01 & 30 & 1e-4 \\
 
 \hline

\multirow{2}*{MultiNLI} & Pretrained & BERT & 20 & 5e-5 & 1 & 0.01 \\ \cline{2-7} & Random & BERT & 20 & 5e-5 & 1 & 0.01 \\

\hline
\end{tabular}
\vspace{0.1in}
\caption{The hyperparameters used to train the models to convergence on each dataset. \\
}
\label{table:hyperparameters}
\end{table}

\begin{table}[h]
\centering
\begin{tabular}{| c | c | c |}
\hline
\textbf{Dataset} & \textbf{Worst-Group Accuracy DRO} & \textbf{Worst-Group Accuracy ERM} \\
\hline
Waterbirds & 84.6 & 39.47 \\
\hline
CelebA Blond/Male & 88.3 & 58.41\\
\hline
MultiNLI & 77.7 & 67.5 \\
\hline
\end{tabular}
\vspace{0.1in}
\caption{Comparison of group DRO summarized from \citet{distributionally} with our own ERM results.}
\label{table:dro}
\end{table}

%% file: parts/6_Framework.tex
\section{Experimental Setup}\label{experimental_setup}

We compare the trends of the average and worst-group train and test performance of different model architectures by varying the model size in terms of depth and width. For CV, we use ResNet \citep{ResNet}, VGG (batchnorm) \citep{vgg}, and MobileNet \citep{sandler2019MobileNetv2}. For the NLP domain, we use BERT \citep{bertoriginal2018}.
Compared to prior work, we use a wide range of commonly-used models  including those run on edge devices to observe the trends in realistic settings.
We also make sure that we follow the experiment settings in prior works.
For example, the setup over the CelebA datasets follows the setup of \citet{DBLP:journals/corr/abs-1912-02292}, varying the width of a ResNet10 model \citep{DBLP:journals/corr/HeZRS15}. The setup over the Waterbirds dataset follows the setup in \citet{mei2020generalization}, training an unregularized logistic regression model over a variable number of projections of the feature representation of the input in a pre-trained ResNet18 model.

\subsection{Datasets}
We look at the trends on four tasks: Waterbirds, CelebA Blond / Male, CelebA Lipstick / Earring, and MultiNLI. The first three are CV tasks whereas MultiNLI is a NLP task. 

\textbf{Waterbirds.} Waterbirds is a synthetic dataset constructed in \citep{distributionally} by cropping out bird photographs from the Caltech-UCSD Birds-200-2011 (CUB) dataset \citep{WelinderEtal2010} and placing them on top of image backgrounds from the Places dataset \citep{zhou2017places}. In Waterbirds, we classify two types of birds: birds that primarily live on land (landbirds) and birds that live on water (waterbirds). These classes are spuriously correlated with the background type: land background or water background. 

Most landbirds are photographed on land and waterbirds on water. Therefore, there are four groups of varying sizes in this dataset. Two large groups of common pairings: landbird on land background, and landbird on water background, and two small groups of less common pairings: waterbird on land background, and waterbird on water background. The background acts as a spurious factor; models typically associate water backgrounds with waterbirds and vice versa. As a result, models tend to fail to generalize to rare groups.

\textbf{CelebA} \citep{liu2015faceattributes} is a large scale multi-feature face dataset with varying backgrounds and poses.  Using the CelebA dataset, we can construct spurious correlations datasets by selecting specific features from the multi-feature dataset. CelebA Blond / Male and CelebA Lipstick/ Earring are two such examples of spurious correlation datasets formed from CelebA.

\textbf{CelebA Blond / Male.} In CelebA Blond / Male, the model classifies images as either containing blond or dark hair. The model classifies images as either containing blond hair or not. The spurious correlation is whether or not the subject is male. 

\textbf{CelebA Lipstick / Earring.} In the CelebA Lipstick / Earring task, the model classifies images as either containing lipstick or no lipstick \citep{khani2020removing}. The spurious correlation is the presence of earrings, which is highly correlated with the presence of lipstick. 

\textbf{MultiNLI.} MultiNLI is a natural language inference dataset introduced in \cite{mnli2018}. The NLI task consists of predicting how a sentence A logically relates to another sentence B. While the three labels (entailment, contradiction, neutral) are represented equally in the dataset, \cite{annotart2018} discovers an annotation artifact: negation words (``nobody", ``no", ``never", and ``nothing"), when present in sentence B, are far more likely to correspond to a contradiction than entailment. Thus, the labels are spuriously correlated with the presence of negation words. 

Table \ref{table:counts} in the appendix describes the number of samples in each group.

\subsection{Models}
We refer to ``pre-trained'' models as those where we finetuned models which were pre-trained (say on ImageNet for CV tasks), or ``trained from scratch'' models where our initial checkpoint begins with randomly initialized weights.

\textbf{VGG (batchnorm).} VGG batchnorm (BN) models are large CNN, which extend on AlexNet using multiple 3x3 sized filters. It was one the of the top performing models for the ImageNet image localization task. We train and test pretrained VGG BN 11, VGG BN 13, VGG BN 16, and VGG BN 18 models.

\textbf{ResNet.} ResNet is a state of the art deep convolutional model that also perform very well on image classification on ImageNet. It was created to minimize the vanishing gradient by adding a residual block that adds weights from previous layers. For the depths, we train and test pretrained ResNet18, ResNet34, ResNet101, and ResNet152 models. For the widths, we train and test pretrained ResNet18 WD4, ResNet18 WD2, ResNet18 W3D4, and ResNet18 models.

\textbf{MobileNet V2 width reductions.} Different width multipliers applied all layers (but the last convolutional layer) of the MobileNetv2 architecture, pretrained on ImageNet \citep{sandler2019MobileNetv2}. We use architectures and pretrained checkpoints from a Github reproduction from \cite{li2019hbonet}.

\textbf{BERT.} We use the BERT architecture for the MultiNLI task, varying the width and depth as in \cite{turc2019}, which showed effectiveness of BERT even with non-standard depths and widths.

\subsection{Training Procedure}\label{training_procedure}
The CV models (ResNet, VGG, MobileNet) were trained with Nvidia GPUs. The BERT models were pretrained and fine-tuned on a TPUv3 through the Google Cloud Platform. For the CV models, we use a batch size of 128, a stochastic gradient descent (SGD) optimizer with a momentum of 0.9, and a step scheduler for the learning rate. For the BERT models, we use a batch size of 64, the Adam optimizer with $\beta_1 = 0.9$ and $\beta_2 = 0.999$, and a linear LR warmup for 10\% of the training epochs followed by linear LR decay to zero. A list of the hyperparameters used can be found in Table \ref{table:hyperparameters}.

\subsubsection{Resampling}
We resample each dataset to verify the consistency of the trends. 95\% confidence interval error bars are included in the graphs for context. Resampling is done by shuffling train, validation, and test data while keeping the proportions of the 4 subgroups in each set.
The models are then trained to convergence and we graph the training and validation errors of the converged models. Within each set of model size experiments (i.e., set of graphs), we trained all the models using the same hyperparameters.

%% file: main.bbl
\begin{thebibliography}{36}
\providecommand{\natexlab}[1]{#1}
\providecommand{\url}[1]{\texttt{#1}}
\expandafter\ifx\csname urlstyle\endcsname\relax
  \providecommand{\doi}[1]{doi: #1}\else
  \providecommand{\doi}{doi: \begingroup \urlstyle{rm}\Url}\fi

\bibitem[Belkin et~al.(2019)Belkin, Hsu, Ma, and Mandal]{belkin2019reconciling}
Mikhail Belkin, Daniel Hsu, Siyuan Ma, and Soumik Mandal.
\newblock Reconciling modern machine learning practice and the bias-variance
  trade-off, 2019.

\bibitem[Bornschein et~al.(2020)Bornschein, Visin, and
  Osindero]{bornschein2020small}
Jorg Bornschein, Francesco Visin, and Simon Osindero.
\newblock Small data, big decisions: Model selection in the small-data regime,
  2020.

\bibitem[Buolamwini \& Gebru(2018)Buolamwini and Gebru]{pmlr-v81-buolamwini18a}
Joy Buolamwini and Timnit Gebru.
\newblock Gender shades: Intersectional accuracy disparities in commercial
  gender classification.
\newblock In Sorelle~A. Friedler and Christo Wilson (eds.), \emph{Proceedings
  of the 1st Conference on Fairness, Accountability and Transparency},
  volume~81 of \emph{Proceedings of Machine Learning Research}, pp.\  77--91.
  PMLR, 23--24 Feb 2018.
\newblock URL \url{https://proceedings.mlr.press/v81/buolamwini18a.html}.

\bibitem[Devlin et~al.(2018)Devlin, Chang, Lee, and
  Toutanova]{bertoriginal2018}
Jacob Devlin, Ming{-}Wei Chang, Kenton Lee, and Kristina Toutanova.
\newblock {BERT:} pre-training of deep bidirectional transformers for language
  understanding.
\newblock \emph{CoRR}, abs/1810.04805, 2018.
\newblock URL \url{http://arxiv.org/abs/1810.04805}.

\bibitem[Goel et~al.(2020)Goel, Gu, Li, and Ré]{goel2020model}
Karan Goel, Albert Gu, Yixuan Li, and Christopher Ré.
\newblock Model patching: Closing the subgroup performance gap with data
  augmentation, 2020.

\bibitem[Gururangan et~al.(2018)Gururangan, Swayamdipta, Levy, Schwartz,
  Bowman, and Smith]{annotart2018}
Suchin Gururangan, Swabha Swayamdipta, Omer Levy, Roy Schwartz, Samuel~R.
  Bowman, and Noah~A. Smith.
\newblock Annotation artifacts in natural language inference data.
\newblock \emph{CoRR}, abs/1803.02324, 2018.
\newblock URL \url{http://arxiv.org/abs/1803.02324}.

\bibitem[Hashimoto et~al.(2018{\natexlab{a}})Hashimoto, Srivastava, Namkoong,
  and Liang]{pmlr-v80-hashimoto18a}
Tatsunori Hashimoto, Megha Srivastava, Hongseok Namkoong, and Percy Liang.
\newblock Fairness without demographics in repeated loss minimization.
\newblock In Jennifer Dy and Andreas Krause (eds.), \emph{Proceedings of the
  35th International Conference on Machine Learning}, volume~80 of
  \emph{Proceedings of Machine Learning Research}, pp.\  1929--1938. PMLR,
  10--15 Jul 2018{\natexlab{a}}.
\newblock URL \url{https://proceedings.mlr.press/v80/hashimoto18a.html}.

\bibitem[Hashimoto et~al.(2018{\natexlab{b}})Hashimoto, Srivastava, Namkoong,
  and Liang]{hashimoto2018fairness}
Tatsunori~B. Hashimoto, Megha Srivastava, Hongseok Namkoong, and Percy Liang.
\newblock Fairness without demographics in repeated loss minimization,
  2018{\natexlab{b}}.

\bibitem[He et~al.(2015{\natexlab{a}})He, Zhang, Ren, and
  Sun]{DBLP:journals/corr/HeZRS15}
Kaiming He, Xiangyu Zhang, Shaoqing Ren, and Jian Sun.
\newblock Deep residual learning for image recognition.
\newblock \emph{CoRR}, abs/1512.03385, 2015{\natexlab{a}}.
\newblock URL \url{http://arxiv.org/abs/1512.03385}.

\bibitem[He et~al.(2015{\natexlab{b}})He, Zhang, Ren, and Sun]{ResNet}
Kaiming He, Xiangyu Zhang, Shaoqing Ren, and Jian Sun.
\newblock Deep residual learning for image recognition.
\newblock \emph{CoRR}, abs/1512.03385, 2015{\natexlab{b}}.
\newblock URL \url{http://arxiv.org/abs/1512.03385}.

\bibitem[Hendrycks et~al.(2021)Hendrycks, Zhao, Basart, Steinhardt, and
  Song]{hendrycks2021natural}
Dan Hendrycks, Kevin Zhao, Steven Basart, Jacob Steinhardt, and Dawn Song.
\newblock Natural adversarial examples, 2021.

\bibitem[Hermann et~al.(2020)Hermann, Chen, and Kornblith]{hermann2020origins}
Katherine~L. Hermann, Ting Chen, and Simon Kornblith.
\newblock The origins and prevalence of texture bias in convolutional neural
  networks, 2020.

\bibitem[Kaushik et~al.(2019)Kaushik, Hovy, and Lipton]{counterfactual2019}
Divyansh Kaushik, Eduard~H. Hovy, and Zachary~C. Lipton.
\newblock Learning the difference that makes a difference with
  counterfactually-augmented data.
\newblock \emph{CoRR}, abs/1909.12434, 2019.
\newblock URL \url{http://arxiv.org/abs/1909.12434}.

\bibitem[Khani \& Liang(2020)Khani and Liang]{khani2020removing}
Fereshte Khani and Percy Liang.
\newblock Removing spurious features can hurt accuracy and affect groups
  disproportionately, 2020.

\bibitem[Li et~al.(2019)Li, Zhou, and Yao]{li2019hbonet}
Duo Li, Aojun Zhou, and Anbang Yao.
\newblock Hbonet: Harmonious bottleneck on two orthogonal dimensions, 2019.

\bibitem[Liu et~al.(2021)Liu, Haghgoo, Chen, Raghunathan, Koh, Sagawa, Liang,
  and Finn]{liu2021just}
Evan~Zheran Liu, Behzad Haghgoo, Annie~S. Chen, Aditi Raghunathan, Pang~Wei
  Koh, Shiori Sagawa, Percy Liang, and Chelsea Finn.
\newblock Just train twice: Improving group robustness without training group
  information, 2021.

\bibitem[Liu et~al.(2015)Liu, Luo, Wang, and Tang]{liu2015faceattributes}
Ziwei Liu, Ping Luo, Xiaogang Wang, and Xiaoou Tang.
\newblock Deep learning face attributes in the wild.
\newblock In \emph{Proceedings of International Conference on Computer Vision
  (ICCV)}, December 2015.

\bibitem[Mei \& Montanari(2020)Mei and Montanari]{mei2020generalization}
Song Mei and Andrea Montanari.
\newblock The generalization error of random features regression: Precise
  asymptotics and double descent curve, 2020.

\bibitem[Menon et~al.(2021)Menon, Rawat, and
  Kumar]{Menon2021OverparameterisationAW}
Aditya~Krishna Menon, Ankit~Singh Rawat, and Sanjiv Kumar.
\newblock Overparameterisation and worst-case generalisation: friend or foe?
\newblock In \emph{ICLR}, 2021.

\bibitem[Nakkiran et~al.(2020)Nakkiran, Kaplun, Bansal, Yang, Barak, and
  Sutskever]{DBLP:journals/corr/abs-1912-02292}
Preetum Nakkiran, Gal Kaplun, Yamini Bansal, Tristan Yang, Boaz Barak, and Ilya
  Sutskever.
\newblock Deep double descent: Where bigger models and more data hurt.
\newblock In \emph{8th International Conference on Learning Representations,
  {ICLR} 2020, Addis Ababa, Ethiopia, April 26-30, 2020}. OpenReview.net, 2020.
\newblock URL \url{https://openreview.net/forum?id=B1g5sA4twr}.

\bibitem[Niven \& Kao(2019)Niven and Kao]{probingbert2019}
Timothy Niven and Hung{-}Yu Kao.
\newblock Probing neural network comprehension of natural language arguments.
\newblock \emph{CoRR}, abs/1907.07355, 2019.
\newblock URL \url{http://arxiv.org/abs/1907.07355}.

\bibitem[Oakden-Rayner et~al.(2020)Oakden-Rayner, Dunnmon, Carneiro, and
  Re]{hidden_strat_clinically_meaningful}
Luke Oakden-Rayner, Jared Dunnmon, Gustavo Carneiro, and Christopher Re.
\newblock Hidden stratification causes clinically meaningful failures in
  machine learning for medical imaging.
\newblock In \emph{Proceedings of the ACM Conference on Health, Inference, and
  Learning}, CHIL '20, pp.\  151–159, New York, NY, USA, 2020. Association
  for Computing Machinery.
\newblock ISBN 9781450370462.
\newblock \doi{10.1145/3368555.3384468}.
\newblock URL \url{https://doi.org/10.1145/3368555.3384468}.

\bibitem[Sagawa et~al.(2020{\natexlab{a}})Sagawa, Koh, Hashimoto, and
  Liang]{distributionally}
Shiori Sagawa, Pang~Wei Koh, Tatsunori~B. Hashimoto, and Percy Liang.
\newblock Distributionally robust neural networks for group shifts: On the
  importance of regularization for worst-case generalization,
  2020{\natexlab{a}}.

\bibitem[Sagawa et~al.(2020{\natexlab{b}})Sagawa, Raghunathan, Koh, and
  Liang]{Sagawa2020AnIO}
Shiori Sagawa, Aditi Raghunathan, Pang~Wei Koh, and Percy Liang.
\newblock An investigation of why overparameterization exacerbates spurious
  correlations.
\newblock In \emph{ICML}, 2020{\natexlab{b}}.

\bibitem[Sandler et~al.(2019)Sandler, Howard, Zhu, Zhmoginov, and
  Chen]{sandler2019MobileNetv2}
Mark Sandler, Andrew Howard, Menglong Zhu, Andrey Zhmoginov, and Liang-Chieh
  Chen.
\newblock Mobilenetv2: Inverted residuals and linear bottlenecks, 2019.

\bibitem[Simonyan \& Zisserman(2014)Simonyan and Zisserman]{vgg}
Karen Simonyan and Andrew Zisserman.
\newblock Very deep convolutional networks for large-scale image recognition.
\newblock \emph{arXiv preprint arXiv:1409.1556}, 2014.

\bibitem[Tu et~al.(2020)Tu, Lalwani, Gella, and He]{tu2020empirical}
Lifu Tu, Garima Lalwani, Spandana Gella, and He~He.
\newblock An empirical study on robustness to spurious correlations using
  pre-trained language models.
\newblock \emph{TACL}, 8, 2020.

\bibitem[Turc et~al.(2019)Turc, Chang, Lee, and Toutanova]{turc2019}
Iulia Turc, Ming-Wei Chang, Kenton Lee, and Kristina Toutanova.
\newblock Well-read students learn better: On the importance of pre-training
  compact models.
\newblock \emph{arXiv preprint arXiv:1908.08962v2}, 2019.

\bibitem[Vapnik(1998)]{vapnik1998statistical}
V~Vapnik.
\newblock Statistical learning theory new york.
\newblock \emph{NY: Wiley}, 1:\penalty0 2, 1998.

\bibitem[Wang et~al.(2021)Wang, Zhou, Sun, and Zhang]{wang2021causal}
Tan Wang, Chang Zhou, Qianru Sun, and Hanwang Zhang.
\newblock Causal attention for unbiased visual recognition, 2021.

\bibitem[Welinder et~al.(2010)Welinder, Branson, Mita, Wah, Schroff, Belongie,
  and Perona]{WelinderEtal2010}
P.~Welinder, S.~Branson, T.~Mita, C.~Wah, F.~Schroff, S.~Belongie, and
  P.~Perona.
\newblock {Caltech-UCSD Birds 200}.
\newblock Technical Report CNS-TR-2010-001, California Institute of Technology,
  2010.

\bibitem[Williams et~al.(2018)Williams, Nangia, and Bowman]{mnli2018}
Adina Williams, Nikita Nangia, and Samuel Bowman.
\newblock A broad-coverage challenge corpus for sentence understanding through
  inference.
\newblock In \emph{Proceedings of the 2018 Conference of the North American
  Chapter of the Association for Computational Linguistics: Human Language
  Technologies, Volume 1 (Long Papers)}, pp.\  1112--1122. Association for
  Computational Linguistics, 2018.
\newblock URL \url{http://aclweb.org/anthology/N18-1101}.

\bibitem[Yang et~al.(2020)Yang, Yu, You, Steinhardt, and
  Ma]{yang2020rethinking}
Zitong Yang, Yaodong Yu, Chong You, Jacob Steinhardt, and Yi~Ma.
\newblock Rethinking bias-variance trade-off for generalization of neural
  networks, 2020.

\bibitem[Zhong et~al.(2021)Zhong, Ghosh, Klein, and
  Steinhardt]{zhong2021larger}
Ruiqi Zhong, Dhruba Ghosh, Dan Klein, and Jacob Steinhardt.
\newblock Are larger pretrained language models uniformly better? comparing
  performance at the instance level, 2021.

\bibitem[Zhou et~al.(2017)Zhou, Lapedriza, Khosla, Oliva, and
  Torralba]{zhou2017places}
Bolei Zhou, Agata Lapedriza, Aditya Khosla, Aude Oliva, and Antonio Torralba.
\newblock Places: A 10 million image database for scene recognition.
\newblock \emph{IEEE Transactions on Pattern Analysis and Machine
  Intelligence}, 2017.

\bibitem[Zhou et~al.(2021)Zhou, Ma, Michel, and Neubig]{zhou2021examining}
Chunting Zhou, Xuezhe Ma, Paul Michel, and Graham Neubig.
\newblock Examining and combating spurious features under distribution shift,
  2021.

\end{thebibliography}
